\documentclass[lettersize,journal]{IEEEtran}
\usepackage{amsmath,amsfonts}
\usepackage{algorithmic}
\usepackage{algorithm}
\usepackage{array}
\usepackage[caption=false,font=normalsize,labelfont=sf,textfont=sf]{subfig}
\usepackage{textcomp}
\usepackage{stfloats}
\usepackage{url}
\usepackage{verbatim}
\usepackage{graphicx}
\usepackage{cite}
\usepackage{times}
\usepackage{epsfig}
\usepackage{amsmath}
\usepackage{amssymb}
\usepackage{tikz}
\usepackage{booktabs}
\usepackage{multirow}
\graphicspath{{./figures/}}
\DeclareGraphicsExtensions{.pdf,.jpeg,.png,.jpg,.eps}
\usepackage{algorithm}
\usepackage{algorithmic}
\usepackage[accsupp]{axessibility}

\begin{document}

\title{High-quality Image Dehazing with Diffusion Model}

\author{Hu Yu, Jie Huang, Kaiwen Zheng and Feng Zhao, Member, IEEE
\thanks{H. Yu, J. Huang, K. Zheng and F. Zhao are affiliated with University of Science and Technology of China, Hefei 230027, China (e-mail:fzhao956@ustc.edu.cn)}
}



\maketitle

\begin{abstract}
Image dehazing is quite challenging in dense-haze scenarios, where quite less original information remains in the hazy image. Though previous methods have made marvelous progress, they still suffer from information loss in content and color in dense-haze scenarios. The recently emerged Denoising Diffusion Probabilistic Model (DDPM) exhibits strong generation ability, showing potential for solving this problem. However, DDPM fails to consider the physics property of dehazing task, limiting its information completion capacity. In this work, we propose DehazeDDPM: A DDPM-based and physics-aware image dehazing framework that applies to complex hazy scenarios. Specifically, DehazeDDPM works in two stages. The former stage physically models the dehazing task with the Atmospheric Scattering Model (ASM), pulling the distribution closer to the clear data and endowing DehazeDDPM with fog-aware ability.
The latter stage exploits the strong generation ability of DDPM to compensate for the haze-induced huge information loss, by working in conjunction with the physical modelling.
Extensive experiments demonstrate that our method attains state-of-the-art performance on both synthetic and real-world hazy datasets. Our code is available at \url{https://github.com/yuhuUSTC/DehazeDDPM}.
\end{abstract}


\section{Introduction}
\IEEEPARstart{H}{aze} is a common atmospheric phenomenon. Images captured in hazy environments usually suffer from information loss both in content and color. The goal of image dehazing is to restore a clean scene from a hazy image. This task has been a longstanding and challenging problem with various applications, such as surveillance systems and autonomous driving, drawing the attention of researchers. As recognized, the haze procedure can be represented by the well-known ASM \cite{mccartney1976optics, narasimhan2002vision}, which is formulated as
\begin{equation}
	\small
		I(x)=J(x) t(x)+A(1-t(x)),
\end{equation}
where $I(x)$ and $J(x)$ denote the hazy image and the clean image respectively, $A$ is the global atmospheric light, and $t(x)$ is the transmission map (we use $trmap$ to denote transmission map here after to avoid confusion). According to ASM, dense-haze corresponds to small transmission map value and less original information.

\begin{figure}[t]
	\begin{center}
		\includegraphics[width=0.95\linewidth]{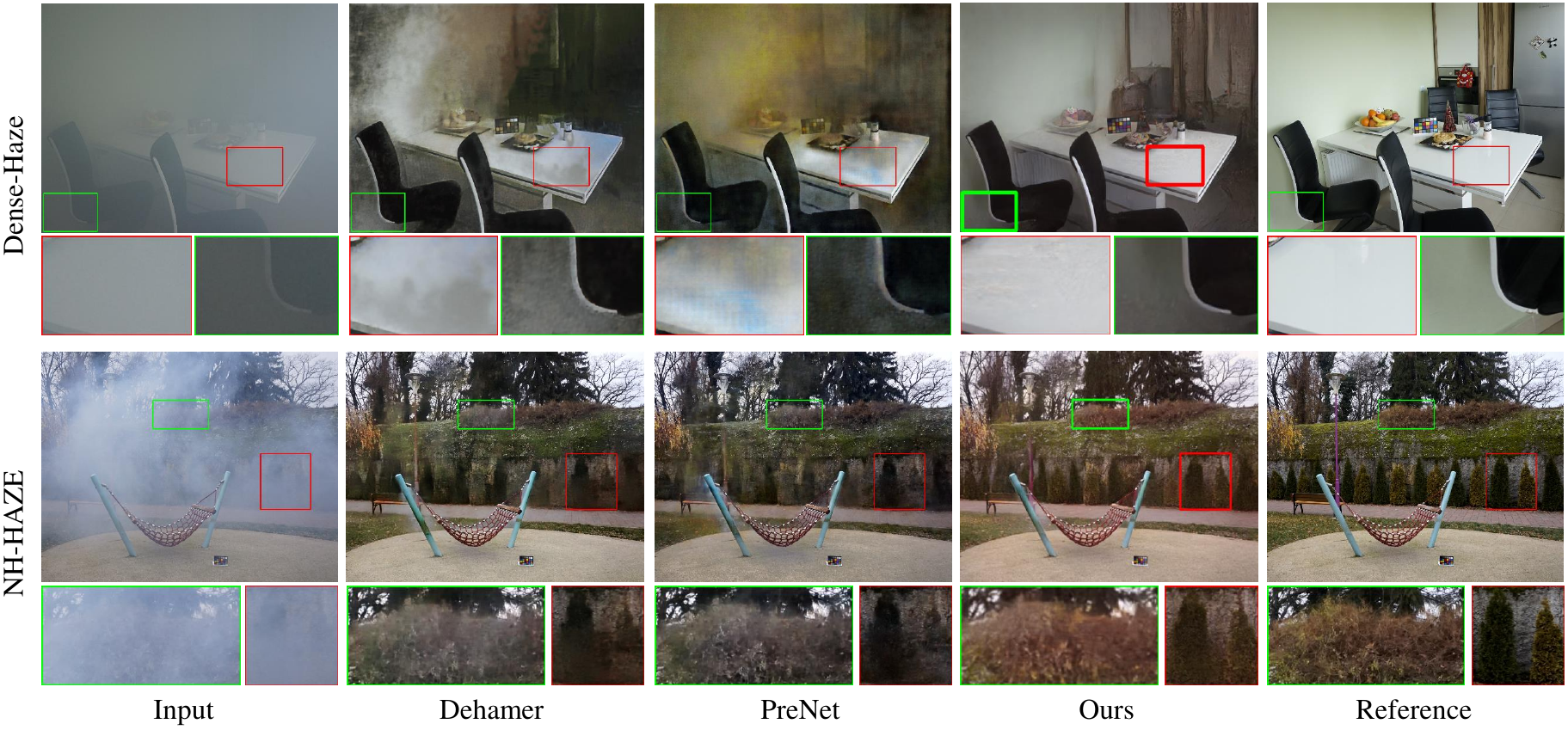}
	\end{center}
	\caption{The visual examples of dehazing results were sampled from real-world hazy images. The second to fourth columns show the results of Dehamer~\cite{guo2022image}, our first-stage, and our DehazeDDPM, respectively. Our method demonstrates unprecedented perceptual quality on the challenging real-world datasets.}
	\label{Tisue}
\end{figure}

With this line, conventional approaches rely on ASM \cite{mccartney1976optics} and adopt priors as external information to estimate the parameters of ASM \cite{berman2016non,he2010single,fattal2014dehazing,fattal2008single}.
However, these hand-crafted image priors are drawn from specific observations with limited robustness, which may not be reliable for modeling the intrinsic characteristics of hazy images.

\begin{figure}[t]
	\begin{center}
		\includegraphics[width=0.95\linewidth]{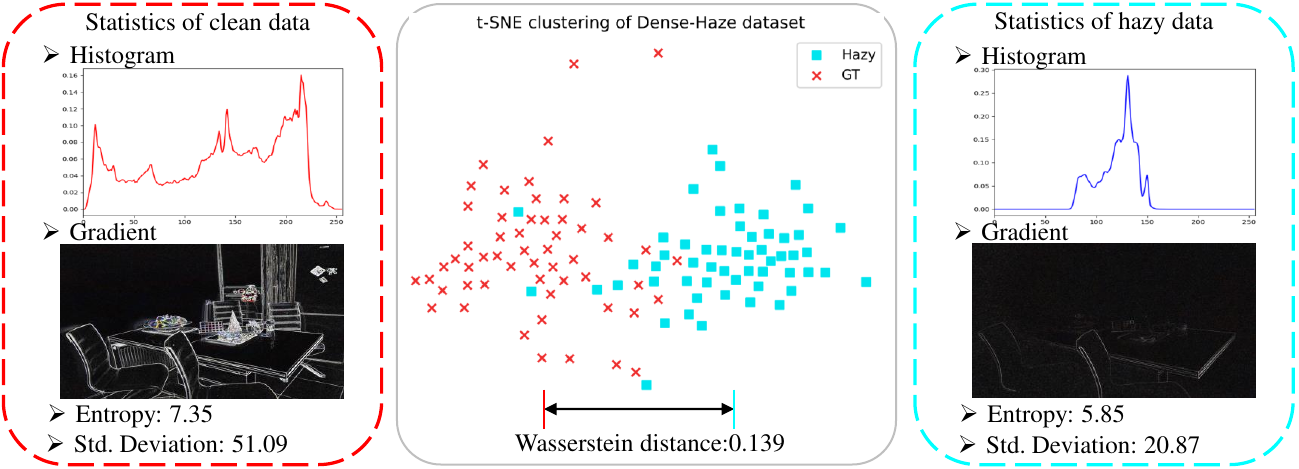}
	\end{center}
	\caption{Statistics illustration of the haze-induced information loss, including t-SNE clustering \cite{van2008visualizing}, distribution distance, histogram, gradient, entropy, and standard deviation.  The dense haze causes massive information loss in content and color.}
	\label{Statistics}
\end{figure}

Inspired by the success of deep learning, numerous approaches \cite{li2017aod,liu2019griddehazenet,guo2019dense,dong2020multi,wu2021contrastive, guo2022image,yufrequency} have been developed to directly learn the hazy to clear mapping.  Though these methods have made marvelous progress, they still suffer from information loss in content and color in dense-haze scenarios, as shown in Fig.~\ref{Tisue}. As for the reason, quite limited original information remains in the hazy image in this challenging case, restricting the information completion ability of such mapping. The detailed statistics illustration of the haze-induced information loss is shown in Fig.~\ref{Statistics}. For example, the t-SNE  \cite{van2008visualizing} as well as the wasserstein distance shows that the distribution of clear and hazy images deviates from each other. Entropy is a measure of the amount of information contained in an image. The average entropy of hazy images is much smaller than that of clear images, which denotes the haze-induced information loss.

Recently, DDPM~\cite{sohl2015deep, song2019generative} has drawn intensive attention due to its strong generation ability. DDPM can produce high-quality images both unconditionally~\cite{ho2020denoising, nichol2021improved, dhariwal2021diffusion} and conditionally~\cite{saharia2022image, whang2022deblurring, rombach2022high}, which pose a new perspective for image dehazing task. However, DDPM fails to consider the physics property of dehazing task, limiting its information completion capacity. For example, the huge distribution difference between dense-haze and clear images makes DDPM struggle with weak and deviate distribution guidance. Besides, DDPM is not aware of the restoration difficulty of different image regions, which is important in modeling the complex distribution in real-world hazy scenarios.

\begin{figure}[t]
	\begin{center}
		\includegraphics[width=0.95\linewidth]{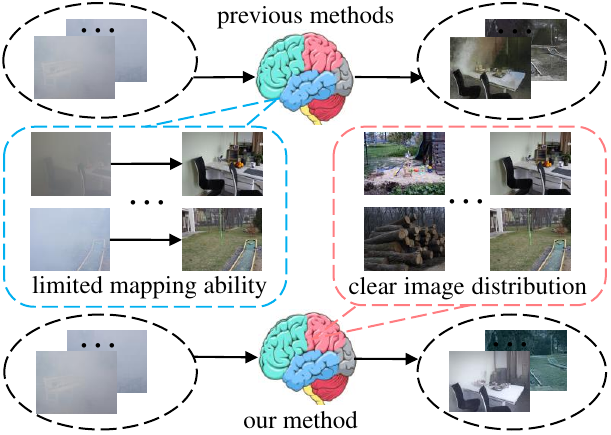}
	\end{center}
	\caption{Thumbnail of main idea. Most previous image dehazing methods learn the mapping from hazy to clear images. Our method memorizes the data distribution of clear images by introducing conditional DDPM into image dehazing.}
	\label{Main_idea}
\end{figure}

In this work, we propose DehazeDDPM: A DDPM-based and physics-aware image dehazing framework that is applicable to complex hazy scenarios. A sketch of the main ideas is shown in Fig.~\ref{Main_idea}. Our DehazeDDPM views image dehazing as a conditional generative modeling task. Instead of learning a mapping, DehazeDDPM memorizes the data distribution of clear images by introducing conditional DDPM into image dehazing, where conditional DDPM approximates the data distribution with appropriate conditions. Therefore, in the challenging dense-haze case, our method largely surpasses previous mapping-based methods.
Besides, the frequency prior of the generation process is leveraged to optimize and constrain the frequency information of the hard region.

Specifically, DehazeDDPM works in two stages. {\bf The former stage} estimates the transmission map $trmap$, the haze-free image $J$, and the atmospheric light $A$ governed by the underlying the Atmospheric Scattering Model (ASM) physics. The estimated haze-free image $J$ has closer distribution to the corresponding clear data than the originally hazy image. The transmission map $trmap$ is exploited as the confidence guidance for the second stage which endows DehazeDDPM with fog-aware ability. {\bf The latter stage} exploits the strong generation ability of DDPM to compensate for the haze-induced huge information loss, by working in conjunction with the physical modelling. The latter stage can recover the details that failed to be retrieved by the first stage, as well as correct artifacts introduced by that stage. 
Besides these, although diffusion model can generate high quality, it spends most of the time to generate high frequency in the whole reverse denoise process. Thus, we impose frequency prior constraints to the training of the diffusion model.
Our method demonstrates unprecedented perceptual quality in image dehazing task, as shown in Fig.~\ref{Tisue}. Extensive experiments demonstrate that our method attains SOTA performance on several image dehazing benchmarks.


Overall, our contributions can be summarized as follows:
\begin{itemize}
	\item We firstly introduce conditional DDPM to tackle the challenging dense-haze image dehazing task by working in conjunction with the physical modelling.
	\item Specifically, physical modelling pulls the distribution of hazy data closer to that of clear data and endows DehazeDDPM with fog-aware ability.
	\item Extensive experiments demonstrate that our method outperforms SOTA approaches on several image dehazing benchmarks with much better FID and LPIPS scores on complex real-world datasets.
\end{itemize}

\section{Related Works}
\label{sec:related}

\noindent
{\bf Image Dehazing.} 
In recent years, we have witnessed significant advances in single image dehazing. Existing methods can be roughly categorized into two classes: physical-based methods and deep learning-based methods.

Physical-based methods depend on the atmospheric scattering model~\cite{mccartney1976optics} and the handcraft priors, such as dark channel prior~\cite{he2010single}, color line prior~\cite{fattal2014dehazing}, color attenuation prior~\cite{zhu2014single}, sparse gradient prior~\cite{chen2016robust}, maximum reflectance prior~\cite{zhang2017fast}, and non-local prior~\cite{berman2016non}. For example, DCP~\cite{he2010single} discovers the dark channel prior to modeling the properties of the hazy-free images, which assumes that the locally lowest intensity in RGB channels should be close to zero in haze-free natural images.
However, the handcraft priors are mainly from empirical observations, which cannot accurately characterize the haze formation process.

Different from the physical-based methods, deep learning-based methods employ convolution neural networks to learn the image prior \cite{cai2016dehazenet,li2017aod,ren2016single,zhang2018densely,liu2018learning,liu2019learning} or directly learn hazy-to-clear translation \cite{li2016underwater, liu2018learning, ren2018gated,liu2019griddehazenet,deng2020hardgan,dong2020multi,qin2020ffa,wu2021contrastive,DehazeYu, liu2022towards, guo2022image, yufrequency, zhou2022fsad,fan2022multiscale, zheng2023curricular, wu2023ridcp}.
For example, AOD-Net~\cite{li2017aod} produces the recovered images by reformulating the Atmospheric Scattering Model. DeHamer~\cite{guo2022image} introduces transformer into image dehazing to combine the global modeling capability of Transformer and the local representation capability of CNN. FSDGN \cite{yufrequency} reveals the relationship between the haze degradation and the characteristics of frequency and jointly explores the information in the frequency and spatial domains for image dehazing.
RIDCP~\cite{wu2023ridcp} presents a paradigm for real image dehazing from the perspectives of synthesizing more realistic hazy data and introducing more robust priors into the network.
Zheng et al.~\cite{zheng2023curricular} proposed a curricular contrastive regularization, which leverages the dehazed results of other existing dehazing methods to provide better lower-bound constraints.

The above techniques have shown outstanding performance on image dehazing. They resort to learn the mapping from hazy images to haze-free counterparts. For example, CNN- and transformer-based methods learn the relationship between different pixels or regions by local or global modeling, respectively.
However, if the image is captured in complex real foggy scenes, the transformation mapping is hard to learn, and the network cannot recover the original clear image via the very limited information in hazy images.
Besides, these approaches share the limitation that they produce a deterministic output, which is at odds with the ill-posedness nature of image dehazing. What's more, the training objectives of minimizing pixel-level distortion are known to be poorly correlated with human perception and often lead to blurry and unrealistic reconstructions ~\cite{whang2022deblurring}, especially in complex real-world hazy scenarios.

Instead of learning a mapping, our method memorizes the information of clear images by introducing conditional diffusion model into image dehazing, where conditional DDPM approximates the data distribution with appropriate conditions. Therefore, in the challenging dense-haze case, our method largely surpasses previous mapping-based methods.

\noindent
{\bf Deep generative models.}
Deep generative models have seen success in learning complex empirical distributions of images and exhibiting convincing image generation results. Generative adversarial networks (GANs), autoregressive models, Normalizing Flows, and variational autoencoders (VAEs) have synthesized striking image samples \cite{van2016conditional, kingma2013auto, kingma2018glow, goodfellow2020generative} and have been applied to conditional tasks such as image dehazing \cite{li2021dehazeflow, deng2020hardgan, dong2020fd, fu2021dw, sharma2020scale, kumar2022orthogonal}.
However, these approaches often suffer from various limitations. For example, GANs capture less diversity than state-of-the-art likelihood-based models \cite{razavi2019generating, nichol2021improved} and require carefully designed regularization and optimization tricks to avoid optimization instability and mode collapse \cite{gulrajani2017improved,miyato2018spectral}.

In contrast, diffusion models, as a class of likelihood-based generative models, possess the desirable properties such as distribution coverage, a stationary training objective, and easy scalability \cite{song2019generative,ho2020denoising,nichol2021improved,dhariwal2021diffusion}. With this line, conditional DDPM \cite{choi2021ilvr, saharia2022image, whang2022deblurring, lugmayr2022repaint, rombach2022high, ozdenizci2022restoring} are developed for image enhancement in low-level vision, such as image super-resolution~\cite{saharia2022image}, image inpainting~\cite{lugmayr2022repaint}, and image deblurring~\cite{whang2022deblurring}. Although DDPM-based methods have been developed for some low-level vision tasks, there is no precedent for usage in image dehazing. Besides, DDPM also fails to consider the physics property of dehazing task, limiting its information completion capacity for hazy images.
Thus, in this paper, we firstly introduce conditional DDPM to tackle the challenging dense-haze image dehazing task by working in conjunction with the physical modelling. 

\section{Preliminaries: DDPM}
\label{sec:Preliminaries}

DDPM is a latent variable model specified by a T-step Markov chain, which approximates a data distribution q(x) with a model $p_\theta (x)$. It contains two processes: the forward diffusion process and the reverse denoise process.

{\bf The forward diffusion process.}
The forward diffusion process starts from a clean data sample $x_0$ and repeatedly injects Gaussian noise according to the transition kernel $q(x_t|x_{t-1})$ as follows:
\begin{equation}
	\small
	\label{1}
	q(x_t|x_{t-1}) = N(x_t;\sqrt{\alpha_t}x_{t-1},(1-\alpha_t)I),  \\
\end{equation}
where $\alpha_t$ can be learned by reparameterization~\cite{kingma2013auto} or held constant as hyper-parameters, controlling the variance of noise added at each step.
From the Gaussian diffusion process, we can derivate closed-form expressions for the marginal1 distribution $q(x_t|x_0)$ and the reverse diffusion step $q(x_{t-1}|x_t,x_0)$ as follows:
\begin{align}
\label{2}
    q(x_t|x_0) = N(x_t;\sqrt{\bar{\alpha} _t}x_0,(1-\bar{\alpha}_t)I), \\
\label{3}
    q(x_{t-1}|x_t,x_0) = N(x_{t-1};\tilde{\mu}_t(x_t,x_0),\tilde{\beta}_tI),
\end{align}
where $\tilde{\mu}_t(x_t,x_0) := \frac{\sqrt{\bar{\alpha}_{t-1}}(1-\alpha_t)}{1-\bar{\alpha}_t}x_0 +  \frac{\sqrt{\alpha_t}(1-\bar{\alpha}_{t-1})}{1-\bar{\alpha}_t}x_t$, $\tilde{\beta}_t:= \frac{1-\bar{\alpha}_{t-1}}{1-\bar{\alpha}_t}(1-\alpha_t)$, and $\bar{\alpha_t} := {\textstyle \prod_{s=1}^{t}} \alpha_s$.

Note that the above-defined forward diffusion formulation has no learnable parameters, and the reverse diffusion step cannot be applied due to having no access to $x_0$ in the inference stage. Therefore, we further introduce the learnable reverse denoise process for estimating $x_0$ from $x_T$.

{\bf The reverse denoise process.}
The DDPM is trained to reverse the process in Equation \ref{1} by learning the denoise network $f_\theta$ in the reverse process. Specifically, the denoise network estimates $f_\theta(x_t,t)$ to replace $x_0$ in Equation \ref{3}.
Note that $f_\theta(x_t,t)$ can predict the Gaussian noise $\varepsilon$ or $x_0$. They deterministicly correspond to each other with Equation \ref{2}.

\begin{equation}
	\small
	\label{4}
	\begin{aligned}
    p_\theta(x_{t-1}|x_t) &= q(x_{t-1}|x_t,f_\theta(x_t,t))\\
    &= N(x_{t-1};\mu _\theta(x_t,t),{\textstyle \sum_{\theta}^{}} (x_t,t)).\\
    \end{aligned}
\end{equation}
\begin{equation}
	\small
	\label{5}
	\mu _\theta(x_t,t) = \tilde{\mu_t}(x_t,x_0),  {\textstyle \sum_{\theta}^{}} (x_t,t) = \tilde{\beta_t}I.
\end{equation}

Similarly, the mean and variance in the reverse Gaussian distribution \ref{4} can be determined by replacing $x_0$ in $\tilde{\mu}_t(x_t,x_0)$ and $\tilde{\beta}_t$ with the learned $\hat{x}_0$

{\bf Training objective and sampling process.}
As mentioned above, $f_\theta(x_t,t)$ is trained to approach the Gaussian noise $\varepsilon$. Thus the final training objective is:
\begin{align}
\label{6}
    L=E_{t,x_0,\varepsilon}\left \| \varepsilon -f_\theta(x_t,t) \right \|_1.
\end{align}

The sampling process in the inference stage is done by running the reverse process. Starting from a pure Gaussian noise $x_T$, we iteratively apply the reverse denoise transition $p_\theta(x_{t-1}|x_t)$ T times, and finally get the clear output $x_0$.

\begin{figure*}[t]
	\begin{center}
		\includegraphics[width=0.95\linewidth]{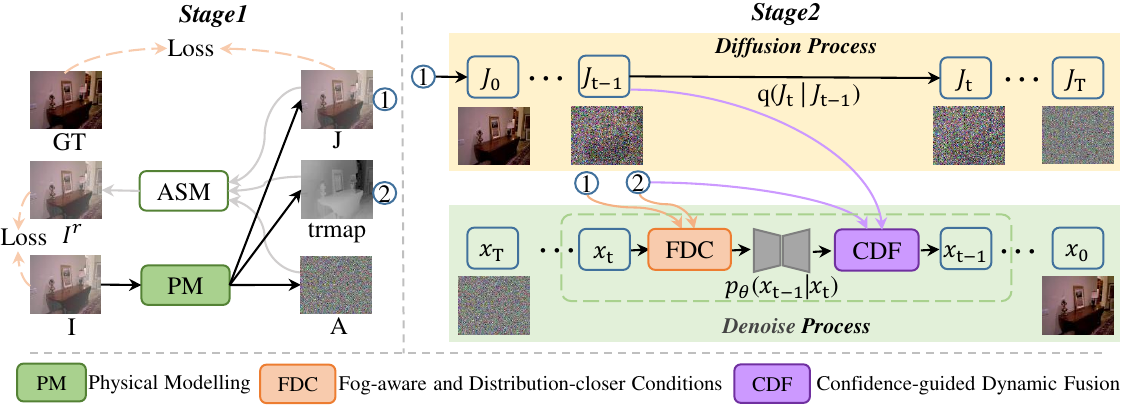}
	\end{center}
	\caption{Overview structure of the proposed DehazeDDPM. DehazeDDPM works in two stages. In the former stage, the physical modelling network generates $J$, $trmap$, and $A$, governed by the underlying SAM physics. For the latter stage, the Fog-aware and Distribution-closer Conditions (FDC) endows the DehazeDDPM with fog-aware ability and pulls the distribution closer to the clear data. The Confidence-guided Dynamic Fusion (CDF) leverages the transmission map as confidence map to incorporate the well restored region of the first stage into the second stage, mitigating the learning difficulty of DDPM for image dehazing.}
	\label{fig_framework}
\end{figure*}

\section{Methodology}
\label{sec:methodology}

The overview structure of our method is presented in Fig.~\ref{fig_framework}. We combine the ASM and DDPM for image dehazing. DehazeDDPM works in two stages. For a hazy image $I$, the former stage firstly outputs the transmission map $trmap$, pseudo haze-free image $J$, and atmospheric light $A$, following the formulation of the ASM. Then, for the latter stage, the learned $trmap$ and $J$ are integrated into the DDPM to pull the distribution of hazy data closer to that of clear data and endow DehazeDDPM with fog-aware ability.

\subsection{Physical modelling.}
The outline of our physics-based network is shown in the left part of Fig.~\ref{fig_framework}. It decomposes the input image $I$ into the $trmap$, $J$, and $A$, governed by the underlying physics. To tease out these components more reliably, we reconstruct the hazy image $I^r$ with the estimated $J$, $trmap$, and $A$ via the formulation of ASM.
\begin{equation}
	\small
		I^r(x)=J(x) trmap(x)+A(1-trmap(x)).
\end{equation}
Then we constrain $J$ and $I^r$ with GT and $I$, respectively.

\begin{figure}[ht]
	\begin{center}
		\includegraphics[width=0.95\linewidth]{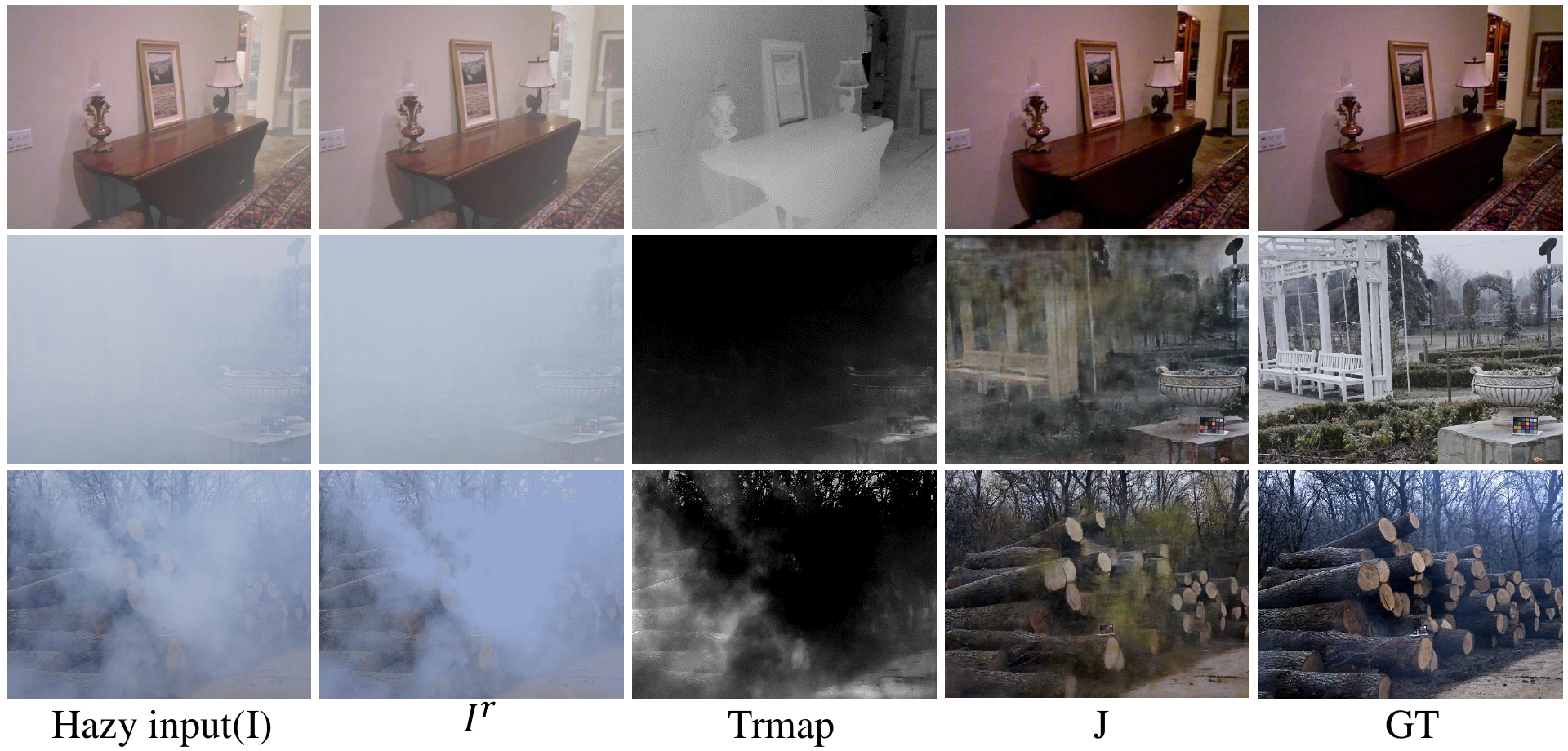}
	\end{center}
	\caption{Different components of the first stage. The three rows are respectively sampled from  SOTS \cite{li2018benchmarking}, Dense-Haze \cite{Dense-Haze_2019}, and NH-HAZE \cite{ancuti2020nh} datasets. The estimated $trmap$ perfectly reflects the density of the fog, where the darker region denotes denser fog.}
	\label{fig_SCATER}
\end{figure}

To demonstrate the effectiveness of physical modelling, we show some examples in Fig.~\ref{fig_SCATER}. As can be seen, the reconstructed $I^r$ is highly similar to the original $I$. Besides, note that the dark region in $trmap$ denotes heavy haze according to the formulation of ASM, thus the estimated $trmap$ is obviously consistent with the haze distribution in hazy input $I$. The dual consistency in reconstructed image $I^r$ and transmission map $trmap$ demonstrates the effectiveness of our physical modelling stage. 

\begin{figure}[ht]
	\begin{center}
		\includegraphics[width=0.95\linewidth]{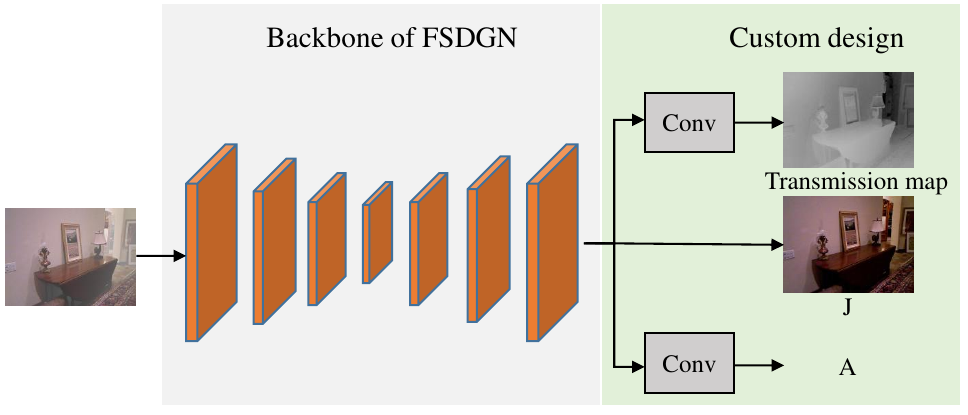}
	\end{center}
    \setlength{\abovecaptionskip}{-0.1cm}
    \setlength{\belowcaptionskip}{-0.2cm}
	\caption{Overview of the network structure of the first stage. Besides the backbone of FSDGN, we make additional design to fit ASM.}
	\label{PreNet}
\end{figure}

About the backbone selection of the first stage. The goal of the first stage is to provide a closer distribution to the clear one and a haze-aware transmission map to guide the diffusion process. Thus, the preferences of the first stage is the best-performed dehazing method by submission. In this paper, we select the FSDGN \cite{yufrequency} as the backbone of the first stage. We make additional design of FSDGN to fit ASM, as shown in Fig.~\ref{PreNet}. Besides, we also employ the classical GridNet \cite{liu2019griddehazenet} as a weaker backbone of the first stage to validate the effectiveness and robustness of our method.

Need to mention that the transmission map in the Atmospheric Scattering Model is correlated with the depth in uniform atmospheric light assumption. But, in complex hazy scenes (For example, the dense or non-homogenous cases), the transmission map is usually not related to the depth but reflects the density of haze, as shown in Fig.~\ref{fig_SCATER}. Some previous dehazing methods \cite{chen2021psd, zheng2023curricular} also employed the ASM in these complex hazy scenes. Besides, slightly different from previous usage, we employ the ASM to get an approximate initialization, instead of the final result. The output transmission map is also capable of working as the dehazing confidence, since the density of haze is strongly correlated with the dehazing difficulty. Thus, it is reasonable to apply ASM in complex hazy scenes as the first stage in our method.

\subsection{Fog-aware and Distribution-closer Conditions.}
Conditional DDPM for low-level vision works by learning to transform a standard normal distribution into an empirical data distribution of clear images with the guidance of degraded images, where the guidance information is denoted as conditions.
Such conditions provide an essential distribution guidance in the T-step stochastic refinement, from standard normal distribution to clear data distribution. This mechanism requires a certain degree of distribution correlation between degraded images and clear images.

Existing DDPM-based image enhancement methods usually directly employ the original degraded image as condition~\cite{lugmayr2022repaint, saharia2022image, ozdenizci2022restoring}. However, such a strategy may not suit dehazing task since the dense-haze images suffer from huge haze-induced information loss in content and color and the distribution of the original hazy input severely deviates from that of the clear image, as shown in Fig.~\ref{Statistics}. This weak condition may fail to provide proper guidance for the learning of clear image distribution. Besides, in real-world complex hazy scenarios, the distribution of fog may be complicated and non-homogeneous. Thus, it is significant to be aware of the distribution of fog for real-world image dehazing task.

\begin{figure}[htbp]
	\begin{center}
		\includegraphics[width=0.95\linewidth]{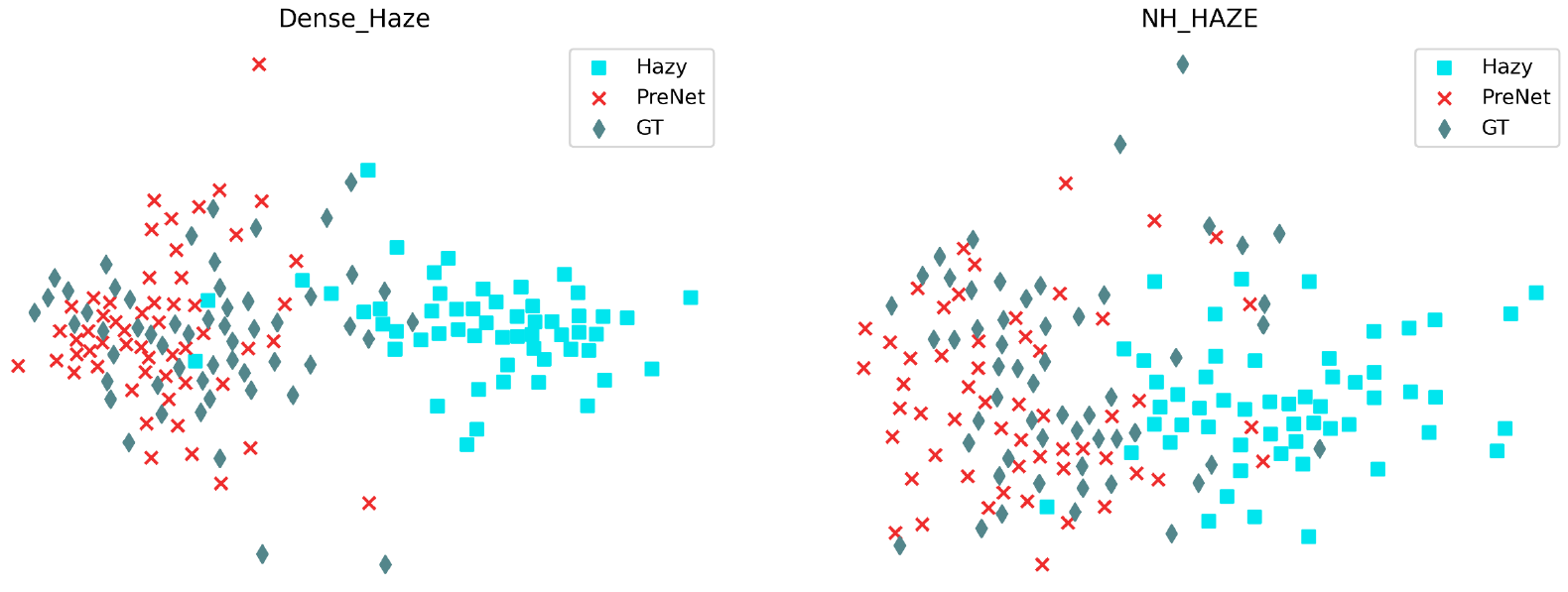}
	\end{center}
	\caption{The t-SNE \cite{van2008visualizing} map of hazy, clear, and the output $J$ of the first stage on Dense-Haze and NH-HAZE datasets. t-SNE is a classic dimensionality reduction and visualization tool. As seen, the outputs of the first stage is closer to the clear images than the hazy inputs, clearly showing that our first stage pulls the distribution of original data closer to the GT. This means that the first stage effectively alleviates the difficulty of DDPM for image dehazing.}
	\label{fig_TSNE}
\end{figure}

To address these problems, we propose the dehazing-customed Fog-aware and Distribution-closer Conditions (FDC). FDC applies $J$ to replace the hazy image $I$ as the condition. As shown in Fig.~\ref{fig_TSNE}, we compare the distribution similarity of the original hazy images, the haze-free images $J$, and the reference clear images on two real-world datasets. Apparently, the distribution of $J$ is more similar to that of the clear than the hazy image $I$. With closer distribution as condition, our method effectively mitigates the difficulty of DDPM for sampling from noise to high-quality images. Besides, we also introduce the transmission map $trmap$ as the condition, indicating explicitly the restoring difficulty of different regions. With this explicit guidance, DehazeDDPM is aware of image-adaptive recovery difficulty and can devote itself to the hard dense-haze region.

Note that Whang et al.~\cite{whang2022deblurring} proposed the "predict and refine" strategy in image deblurring task. However, this plain strategy is not applicable to dehazing task, as it did not take the different restoration difficulties of hazy images and the instincts of dehazing task into consideration. The "predict" stage performs terribly on real-world hazy images. Thus the corresponding residual is severely distorted and is even harder to learn and converge than the original hazy image. Ablation experiment label b resembles the "predict and refine" setting and its performance drops dramatically. In contrast, our method is free of this trouble, as our conditional DDPM in the second stage directly memorizes the clear data distribution with dehazing-customed conditions.



\subsection{Confidence-guided Dynamic Fusion.}
\begin{figure}[t]
	\begin{center}
		\includegraphics[width=0.95\linewidth]{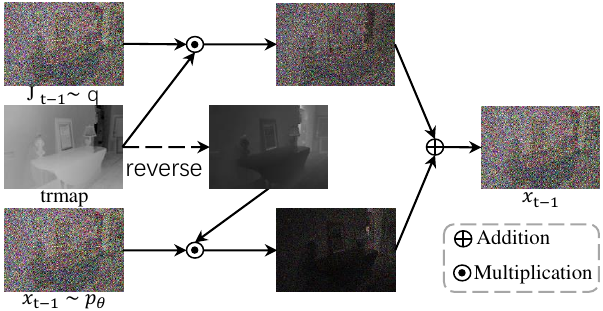}
	\end{center}
	\caption{Confidence-guided Dynamic Fusion. The transmission map from the first stage works as the confidence map guiding the learning of the second stage. Since the first stage performs relatively well in the "easy" region of the hazy image, we leverage this well restored information with confidence-guided dynamic fusion module to mitigate the learning difficulty of DDPM for image dehazing.}
	\label{fig_Fusion}
\end{figure}

For the DDPM refinement stage, besides integrating the dehazed image $J$ and transmission map $trmap$ into the condition phase, these two priors are also introduced to the denoise stage to adaptively excavate the well-dehazed part of the dehazed image $J$ for finer results.
Concretely, given a hazy image as input, the first stage predicts $J$. (1) For synthetic hazy images in SOTS datasets, the first stage behaves pretty well and produces high-fidelity dehazed images $J$ that look perceptually close to the reference ground truths. (2) For the complex real-world hazy images in Dense-Haze and NH-HAZE datasets, the first stage fails to remove haze and suffer from severe color distortion in the dense haze region, but it usually can remove the haze in the thin fog region, as shown in Fig.~\ref{Tisue}.
To better exploit the dehazing ability of existing methods on synthetic hazy images and thin fog regions of real-world hazy images, we propose the Confidence-guided Dynamic Fusion (CDF) to adaptively fuse the output of the first stage and DDPM during the denoise process, as shown in Fig.~\ref{fig_Fusion}. CDF helps adaptively excavate the best of the two stages, and is particularly helpful for performance lift on synthetic hazy dataset.
Specifically, we get $J_{t-1}$ from $J$ via Equation \ref{2} in the diffusion process, and get $x_{t-1}$ from $x_{T}$ via Equation \ref{3} in the reverse denoise process. Then, the final $x_{t-1}$ is obtained as follows.
\begin{equation}
	\small
	\label{7}
	x_{t-1} = trmap \cdot  J_{t-1} + (1-trmap) \cdot  x_{t-1}.
\end{equation}

We conduct the above-mentioned fusion at selective step $t\in [1,T]$, not every step in $T$ timestep. The selection of fusion step $t$ is dependent on the restoration difficulty of the hazy images. We empirically and experimentally find that a severe-distorted original hazy image $I$ corresponds to early sampling fusion, i.e., large step $t$. In our experiment, we choose small step $t$ for SOTS dataset and large step $t$ for Dense-Haze and NH-HAZE datasets.
A pseudocode for the proposed DehazeDDPM is shown in Algorithm \ref{alg:DehazeDDPM}.

\begin{algorithm}[ht]
\caption{DehazeDDPM.}
\label{alg:DehazeDDPM}
\begin{algorithmic}[1]
    \REQUIRE $f_\theta$: Denoiser network, $g_\theta$: the first stage, \\
    $I$: Hazy input image, $\alpha_{1:T}$: Noise schedule.
    \STATE $J, trmap, A \gets g_\theta(I)$ 
    \STATE $z_T \sim N(0, I)$    
    \FOR{$t = T, \ldots, 1$}
        \STATE $\varepsilon \sim N(0, \ I)$
        \STATE $x_{t-1} \gets \tilde{\mu}_t(x_t, f_\theta(x_t, J, trmap, t)) + \tilde{\beta}_t \varepsilon $ 
        \STATE $J_{t-1} \gets \sqrt{\bar{\alpha} _{t-1}}J + (1-\bar{\alpha}_{t-1})\varepsilon $
        \STATE $x_{t-1} \gets trmap \cdot  J_{t-1} + (1-trmap) \cdot  x_{t-1} $
    \ENDFOR
    \RETURN $x_0$
\end{algorithmic}
\end{algorithm}

\subsection{Frequency prior of the generation process.}
\begin{figure}[h]
	\begin{center}
		\includegraphics[width=0.95\linewidth]{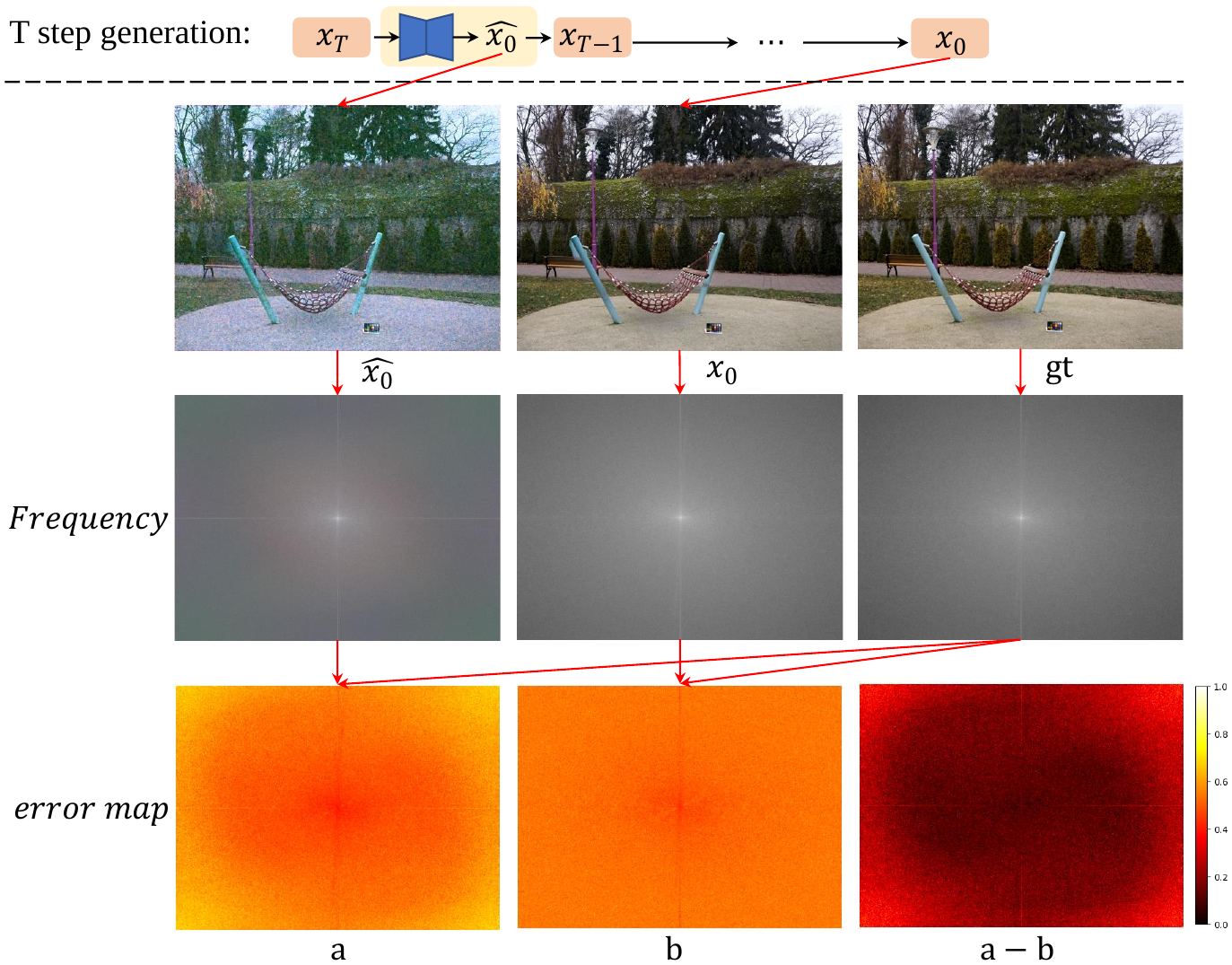}
	\end{center}
	\caption{Frequency prior of the generation process.}
	\label{Frequency}
\end{figure}

Maximum-likelihood training spends a disproportionate amount of capacity on modeling the perceptible, high-frequency details of images~\cite{rombach2022high}. We further excavate the frequency prior of the conditional generation process. As shown in Fig.~\ref{Frequency}, the first row is the intermediate results of the T-steps reverse process. And the second row is the frequency of the corresponding results. The third row is the error maps between frequency maps. 
We can draw two conclusions from our observation.
(1) The diffusion model well recovers most of the color, illumination and structure information in the first reverse step of the whole T-steps reverse process (refer to $\hat{x_0}$ in the first reverse step). And the difference between $\hat{x_0}$ and gt mainly lies in high frequency part (refer to error map $a$). (2) The reverse generation process speeds most of the steps reconstructing the high frequency information. For example, the error map $a-b$ indicates that the final result $x_0$ has closer high frequency part to gt than the initial result $\hat{x_0}$.

Based on such frequency prior hidden in the generation process, we correspondingly devise a frequency prior optimization strategy to help better recover the high frequency information. Specifically, apart from the commonly acknowledged optimization with Equation \ref{6}, we add an additional frequency prior loss as follows:
\begin{equation}
	\small
	\label{loss}
	L_{Fre}= W \left \| Fre(x) - Fre(g(f_\theta(x_t, J, trmap, t))) \right \|_1,
\end{equation}
where $g()$ denotes the transformation from noise to $x_0$ using Equation \ref{2}. $Fre()$ denotes the Fourier transform to get the frequency map. $W$ is the adaptive weight to allocate relatively higher factor to high frequency part to promote the learning of high frequency. $W$ is implemented in the form of $W(x,y) = [(x-W/2)^2 + (y-H/2)^2]/[(W/2)^2 + (H/2)^2]$. Finally, $L_{Fre}$ is added to Equation \ref{6} with a weight of 0.01.

\subsection{Color shift problem with diffusion model.}
Existing methods~\cite{song2020improved, deck2023easing} reveal that generated images of score-based models can suffer from errors in their spatial means, an effect, referred to as a color shift, which grows for larger images. This may arise from the high variance of the model with respect to the task of predicting the spatial mean of the score function. For the conditional diffusion model applied in our method, the color shift problem is largely alleviated with the pre-dehazed image from the first stage as distribution guidance. While, we still observe slight color shift in some generated samples. To further get rid of color shift in image dehazing, we adopt the exponential moving average (EMA) strategy during training, following ~\cite{song2020improved}.

\section{Experiments}
\label{sec:experiments}
In this section, we firstly introduce implement details of our experiment. Then, we make a comprehensive comparison with existing methods quantitatively and visually. Experiments on the challenging Dense-Haze \cite{Dense-Haze_2019} and NH-HAZE \cite{ancuti2020nh} datasets as well as the common synthetic dataset RESIDE~\cite{li2018benchmarking} demonstrate the superiority of our method. We also validate the superior generalization ability of our methods on URHI and RTTS datasets~\cite{li2018benchmarking}, which are hazy natural images. Furthermore, extensive ablation studies are conducted to justify the design rationality of our method.

\subsection{Experiment Setup}

\noindent
{\bf Datasets.}
We train and evaluate our models on both synthetic and real-world image dehazing datasets. For real-world challenging scenes, we adopt two real-world datasets: Dense-Haze~\cite{Dense-Haze_2019} and NH-HAZE~\cite{ancuti2020nh}, to evaluate the performance of our method.
Dense-Haze consists of dense and homogeneous hazy scenes, and NH-HAZE consists of nonhomogeneous hazy scenes. Both of these two datasets consist of $55$ paired images. We choose the former $50$ paired images as training set, and the latter $5$ as testing set.
We also employ RESIDE \cite{li2018benchmarking} dataset for validating the performance in synthetic scenes. The subset Indoor Training Set (ITS) of RESIDE contains a total of $13990$ hazy indoor images, generated from $1399$ clear images. The subset Synthetic Objective Testing Set (SOTS) of RESIDE consists of $500$ indoor hazy images and $500$ outdoor ones. We apply ITS and SOTS indoor as our training and testing sets. We also employ the natural hazy datasets, URHI and RTTS, to evaluate the generalization ability of our method. 

\begin{table*}[t]
\small
\setlength{\tabcolsep}{4pt}
\caption{Quantitative image dehazing results on the Dense-Haze \cite{Dense-Haze_2019}, NH-HAZE \cite{ancuti2020nh}, and SOTS \cite{li2018benchmarking} datasets. 
\colorbox{blue!15}{Best values} and \colorbox{green!15}{second-best values} for each each metric are color-coded. $\downarrow$ denote lower is better and $\uparrow$ means higher is better.} 
\label{performance_Comparison}
\centering
\begin{tabular}{lcccccccccccc}
	\toprule
	\multirow{3}{*}{Method}             & \multicolumn{4}{c}{Dense-Haze \cite{Dense-Haze_2019}}             
	& \multicolumn{4}{c}{NH-HAZE \cite{ancuti2020nh}}        & \multicolumn{4}{c}{SOTS \cite{li2018benchmarking}}   \\ 
	
	\cmidrule(lr){2-5} \cmidrule(lr){6-9} \cmidrule(lr){10-13}
	& \multicolumn{2}{c}{\textbf{Perceptual}}& \multicolumn{2}{c}{\textbf{Distortion}} & \multicolumn{2}{c}{\textbf{Perceptual}}& \multicolumn{2}{c}{\textbf{Distortion}} & \multicolumn{2}{c}{\textbf{Perceptual}}& \multicolumn{2}{c}{\textbf{Distortion}}\\
	
	\cmidrule(lr){2-3} \cmidrule(lr){4-5} \cmidrule(lr){6-7} \cmidrule(lr){8-9} \cmidrule(lr){10-11} \cmidrule(lr){12-13}
	& FID$\downarrow$ & LPIPS$\downarrow$  & PSNR$\uparrow$ & SSIM$\uparrow$   & FID$\downarrow$ & LPIPS$\downarrow$  & PSNR$\uparrow$ & SSIM$\uparrow$  & FID$\downarrow$ & LPIPS$\downarrow$  & PSNR$\uparrow$ & SSIM$\uparrow$ \\
	\midrule
	DCP \cite{he2010single}              &343.65 & 0.6050 & 10.06 & 0.3856       & 348.36 & 0.3994 & 10.57 & 0.5196     
	& 30.85 & 0.0694 & 15.09 & 0.7649       \\
	AOD-Net \cite{li2017aod}             & 415.82 & 0.5991 & 13.14 & 0.4144      & 461.42 & 0.4947 & 15.40 & 0.5693         & 48.56 & 0.0988 & 19.82 & 0.8178       \\
	GridNet \cite{liu2019griddehazenet}  & 429.73 & 0.5102 & 13.31 & 0.3681      & 331.15 & 0.3046 & 13.80 & 0.5370         & 3.93 & 0.0081 & 32.16 & 0.9836        \\
	FFA-Net \cite{qin2020ffa}            & 413.22 & 0.4976 & 14.39 & 0.4524      & 374.31 & 0.3653 & 19.87 & 0.6915         & 2.30 & 0.0048 & 36.39 & 0.9886        \\
	MSBDN \cite{dong2020multi}           & 335.03 & 0.5358 & 15.37 & 0.4858      & 287.81 & 0.2918 & 19.23 & 0.7056         & 8.13 & 0.0288 & 33.79 & 0.9840        \\
	AECR-Net \cite{wu2021contrastive}    & 335.18 & 0.5368 & 15.80 & 0.4660      & 196.38 & 0.2782 & 19.88 & 0.7073        & 3.18 & 0.0068 & 37.17 & 0.9901        \\
	Dehamer \cite{guo2022image}          & 223.65 & 0.4796 & 16.62 & 0.5602      & \colorbox{green!15}{138.49} & 0.2296 & \colorbox{green!15}{20.66} & 0.6844         & 2.54 & 0.0046 & 36.63 & 0.9881        \\
	FSDGN \cite{yufrequency}       & \colorbox{green!15}{202.94} & \colorbox{green!15}{0.4190} & \colorbox{green!15}{16.91} & \colorbox{green!15}{0.5806}      & 177.63 & \colorbox{green!15}{0.2248} & 19.99 & \colorbox{green!15}{0.7106}         & \colorbox{blue!15}{1.54} & \colorbox{blue!15}{0.0037} & \colorbox{blue!15}{38.63}   &\colorbox{blue!15}{0.9903}    \\ 
	RIDCP \cite{wu2023ridcp}          & 316.38 & 0.5507 & 8.0857 & 0.4173      & 367.67 & 0.3578 & 12.27 & 0.4996         & 45.24 & 0.6506 & 18.3562 & 0.7526        \\

	\cmidrule{1-13}
	First-stage                         & 221.17 & 0.4272 & 15.86 & 0.5588      & 184.30 & 0.2376 & 18.73 & 0.6494         & 1.72 & 0.0043 & 37.88  & 0.9887        \\ 
	DehazeDDPM     &\colorbox{blue!15}{171.06} &\colorbox{blue!15}{0.2994} &\colorbox{blue!15}{19.04}  &\colorbox{blue!15}{0.5922}
	& \colorbox{blue!15}{121.06} & \colorbox{blue!15}{0.1624} & \colorbox{blue!15}{22.28}  & \colorbox{blue!15}{0.7309}  
	& \colorbox{green!15}{1.68} & \colorbox{green!15}{0.0041} & \colorbox{green!15}{38.03}   &\colorbox{green!15}{0.9895}  \\

	\bottomrule
\end{tabular}
\end{table*}

\begin{figure*}[h]
	\begin{center}
		\includegraphics[width=0.985\linewidth]{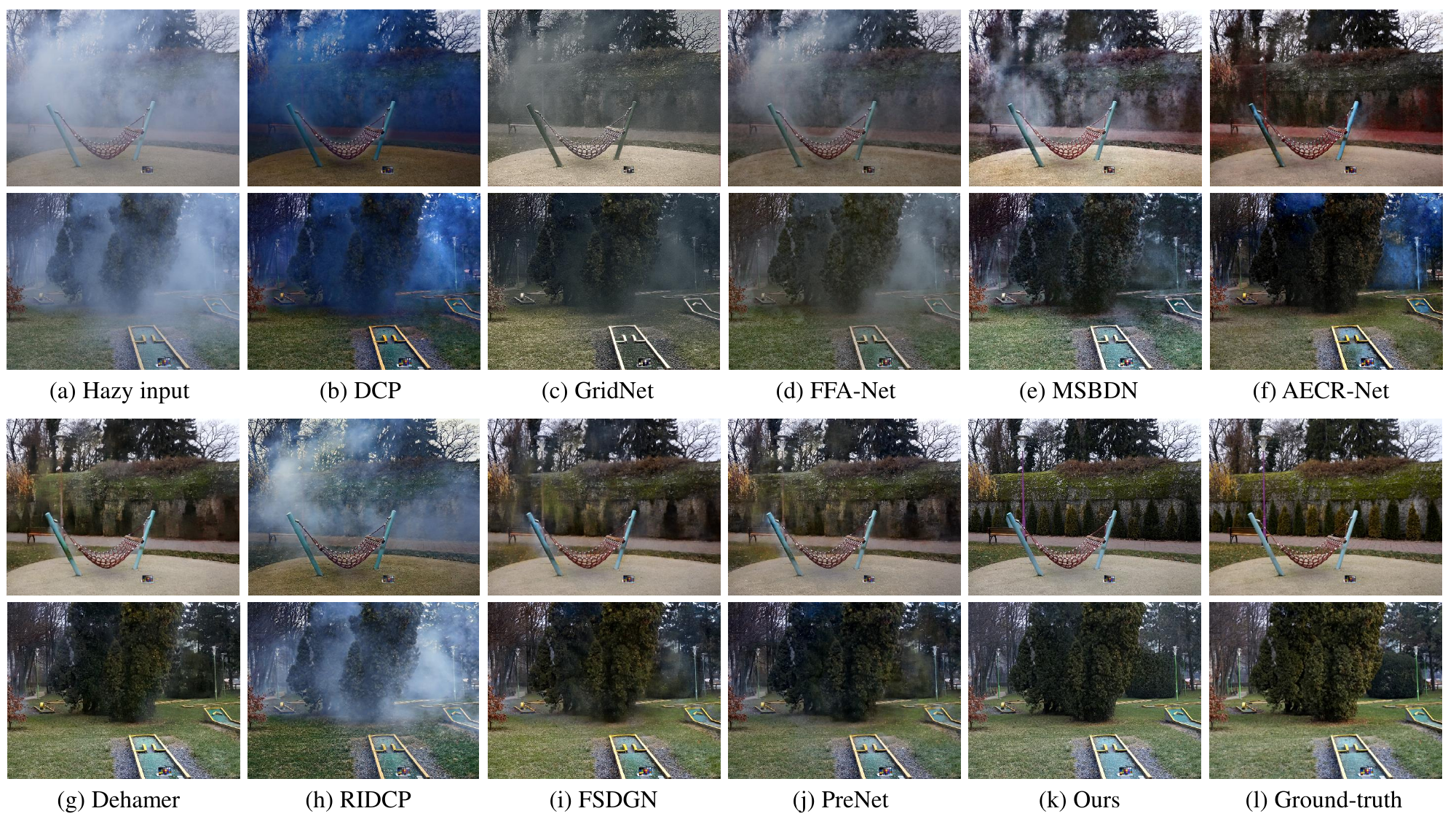}
	\end{center}
	\caption{Comparison of visual results on NH-HAZE \cite{ancuti2020nh} dataset. Zoom in for best view.}
	\label{Visualization_NH}
\end{figure*}

\noindent
{\bf Implementation Details.}
We pretrain the physical modelling network with the backbone in FSDGN \cite{yufrequency}. For the training of the first stage, we use ADAM as the optimizers with $\beta_{1}=0.9$, and $\beta_{2}=0.999$, and the initial learning rate is set to $2\times10^{-4}$. The learning rate is adjusted by the cosine annealing strategy~\cite{he2019bag}. The batch and patch sizes are set to $16$ and $256\times256$, respectively.
We use the Adam optimizer with a linear warmup schedule over 10k training steps, followed by a fixed learning rate of 1e-4 for DehazeDDPM network.
We use 92.6M parameters for our DehazeDDPM network, and 2.8M for the first stage. All the experiments are conducted on a NVIDIA V100 32G.

\noindent
{\bf Evaluation Metrics.} 
We evaluate our method on six different metrics: FID~\cite{heusel2017gans}, LPIPS~\cite{zhang2018unreasonable}, PSNR, SSIM~\cite{wang2004image}, BRISQUE \cite{mittal2012no}, and NIQE \cite{mittal2012making},
in which FID and LPIPS are perceptual-based metrics, PSNR and SSIM are distortion-based metrics, BRISQUE and NIQE are two well-known no-reference image quality assessment indicators.
The commonly adopted image quality scores, PSNR and SSIM, do not reflect human preference well. 
Thus, we also resort to perceptual-based metrics, FID and LPIPS, to compare the quality of image dehazing methods.
Besides, we employ BRISQUE and NIQE to quantitatively assess the performance on the unlabeled URHI and RTTS dataset.

\noindent
{\bf Comparison Methods.} 
We compare our method with the SOTA methods qualitatively and quantitatively, including one prior-based algorithm (DCP~\cite{he2010single}), seven deep learning-based methods (GridNet~\cite{liu2019griddehazenet}, FFA-Net~\cite{qin2020ffa}, MSBDN~\cite{dong2020multi}, AECR-Net~\cite{wu2021contrastive}, Dehamer \cite{guo2022image}, RIDCP \cite{wu2023ridcp}, and FSDGN~\cite{yufrequency}), and our pretrained first-stage.

\noindent
{\bf Difference between the first stage and FSDGN backbone.} 
Most existing dehazng methods only generates the dehazed image as output, including the backbone methods we select for the first stage. While the first stage
is designed to generates several outputs in the ASM fashion. Thus, we slightly modify and retrain the network of the backbone methods to fit the first stage. This explains the slight performance difference between the FSDGN and the first stage in Table~\ref{performance_Comparison}.

\begin{figure*}[ht]
	\begin{center}
		\includegraphics[width=0.985\linewidth]{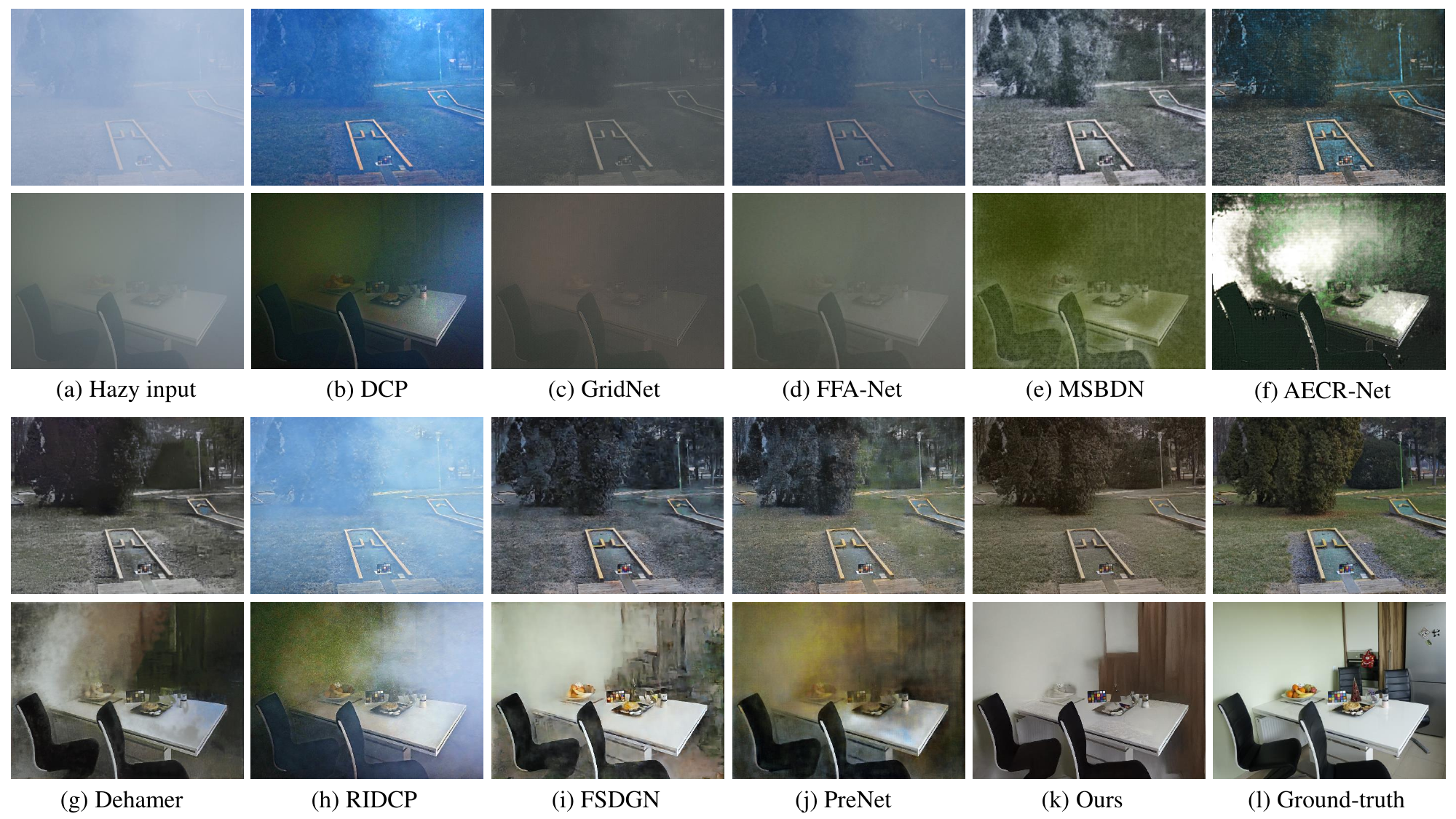}
	\end{center}
	\caption{Comparison of visual results on Dense-Haze \cite{Dense-Haze_2019} dataset. Zoom in for best view.}
	\label{Visualization_Dense}
\end{figure*}

\begin{figure*}[t]
	\begin{center}
		\includegraphics[width=0.985\linewidth]{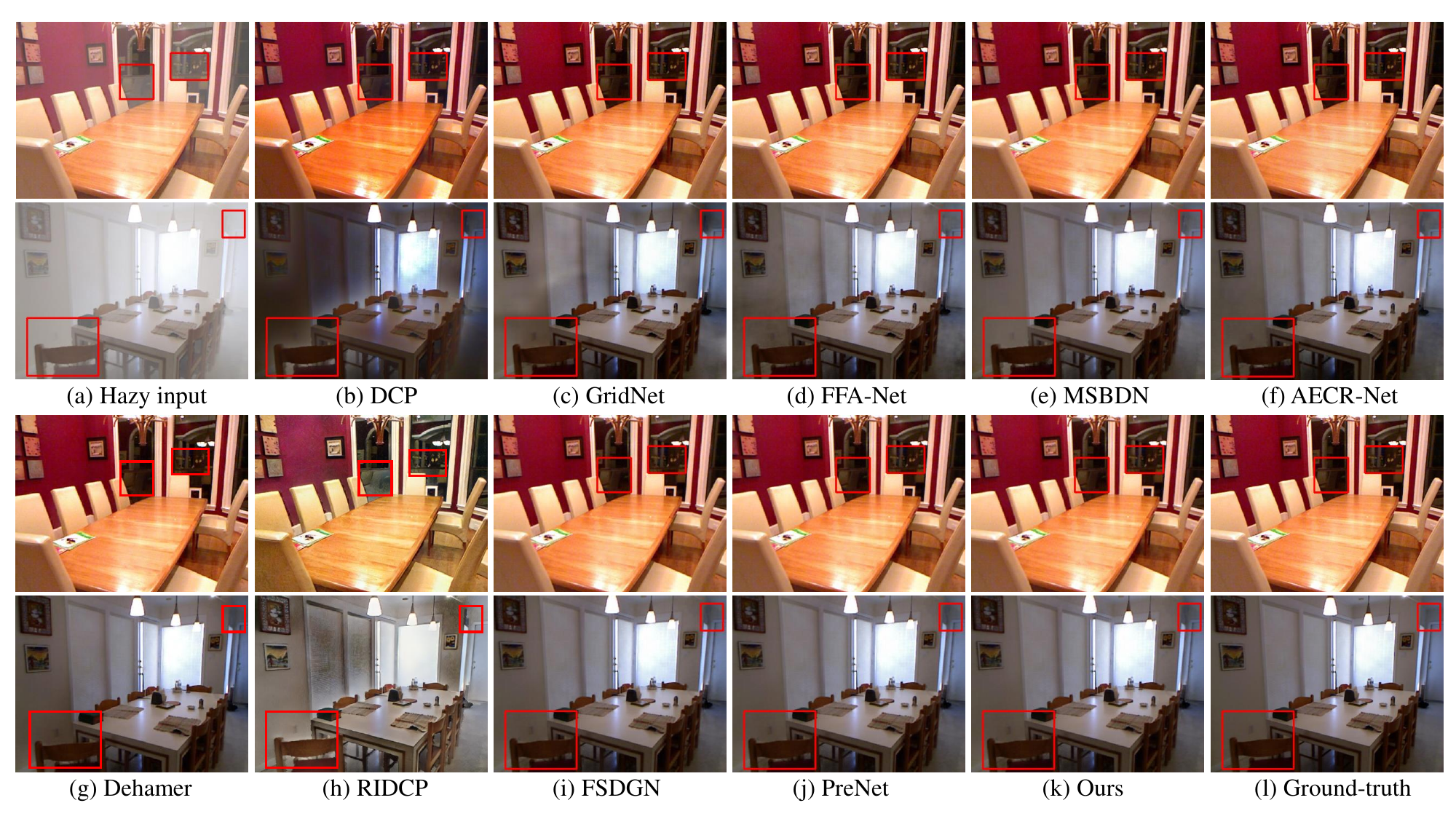}
	\end{center}
	\caption{Comparison of visual results on SOTS \cite{li2018benchmarking} dataset. Red boxes indicate the obvious differences. Zoom in for best view.}
	\label{Visualization_SOTS}
\end{figure*}

\subsection{Comparison with State-of-the-art Methods.}
\noindent
{\bf Comparison with SOTAs on real-world hazy images} 
Table~\ref{performance_Comparison} compares the quantitative results of different methods on Dense-Haze and NH-HAZE datasets. On these two datasets, our method achieves the best performance with both perceptual and distortion metrics. For PSNR scores, our method largely exceeds existing SOTAs by 2-3 dB. For perceptual metric FID, our method also largely surpasses existing methods. The results on such challenging datasets effectively demonstrate the advantages of our approach.

We also show the visual comparison with other techniques on real hazy images sampled from Dense-Haze and NH-HAZE testing sets. The visual results are presented in Fig.~\ref{Visualization_Dense} and Fig.~\ref{Visualization_NH}, respectively. These two datasets are extremely challenging. Dense-Haze image is blocked by thick and homogeneous fog, where the structure and color information are severely distorted. Besides, NH-HAZE image is captured in nonhomogeneous hazy scene. As shown in Fig.~\ref{Visualization_Dense}, existing methods fail to remove the dense haze and suffer from huge information loss in content and color. In contrast, our method can remove nearly all the haze, effectively completes the lost information, and generates the highest-fidelity dehazed results that also look perceptually close to the reference ground truths.
For the results in Fig.~\ref{Visualization_NH}, Dehamer and FSDGN produce relatively pleasing results, however, artifacts and blurs still emerge in the hazy region. In contrast, our method looks more visually pleasing than the compared results, and demonstrates unprecedented perceptual quality in image dehazing task

\noindent
{\bf Comparison on synthetic hazy images.}
Table~\ref{performance_Comparison} compares the quantitative results of different methods on SOTS dataset. We also show visual comparison on the typical hazy images sampled from the SOTS dataset in Fig.~\ref{Visualization_SOTS}. Our method outperforms most of the existing methods. Note that due to the synthetic thin-fog data property of SOTS dataset, existing methods already demonstrate pretty high performance and visual pleasing dehazed results on this dataset. For example, the visual effects of different methods in Fig.~\ref{Visualization_SOTS} is hard to distinguish. Additional performance lift is weak in perception, and may even lead to overfitting. Besides, the performance of our method can be further improved with a better pretrained first-stage network.

\subsection{More experiments and analyses.}

\noindent
{\bf Comparison on natural hazy images (generalization ability).}
We employ the model pretrained on NH-HAZE dataset to evaluate the generalization ability of our method on natural hazy images. Firstly, we need to emphasis that DehazeDDPM is trained on small-scale NH-HAZE dataset. Thus, for the generalization validation on the large-scale URHI and RTTS datasets \cite{li2018benchmarking}, we cannot expect the dehazing results to be as convincing as on NH-HAZE dataset. A reasonable and fair way is to compare generalization performance with existing dehazing methods.
 
\begin{figure}[h]
	\centering
	\includegraphics[width=\linewidth,scale=1.00]{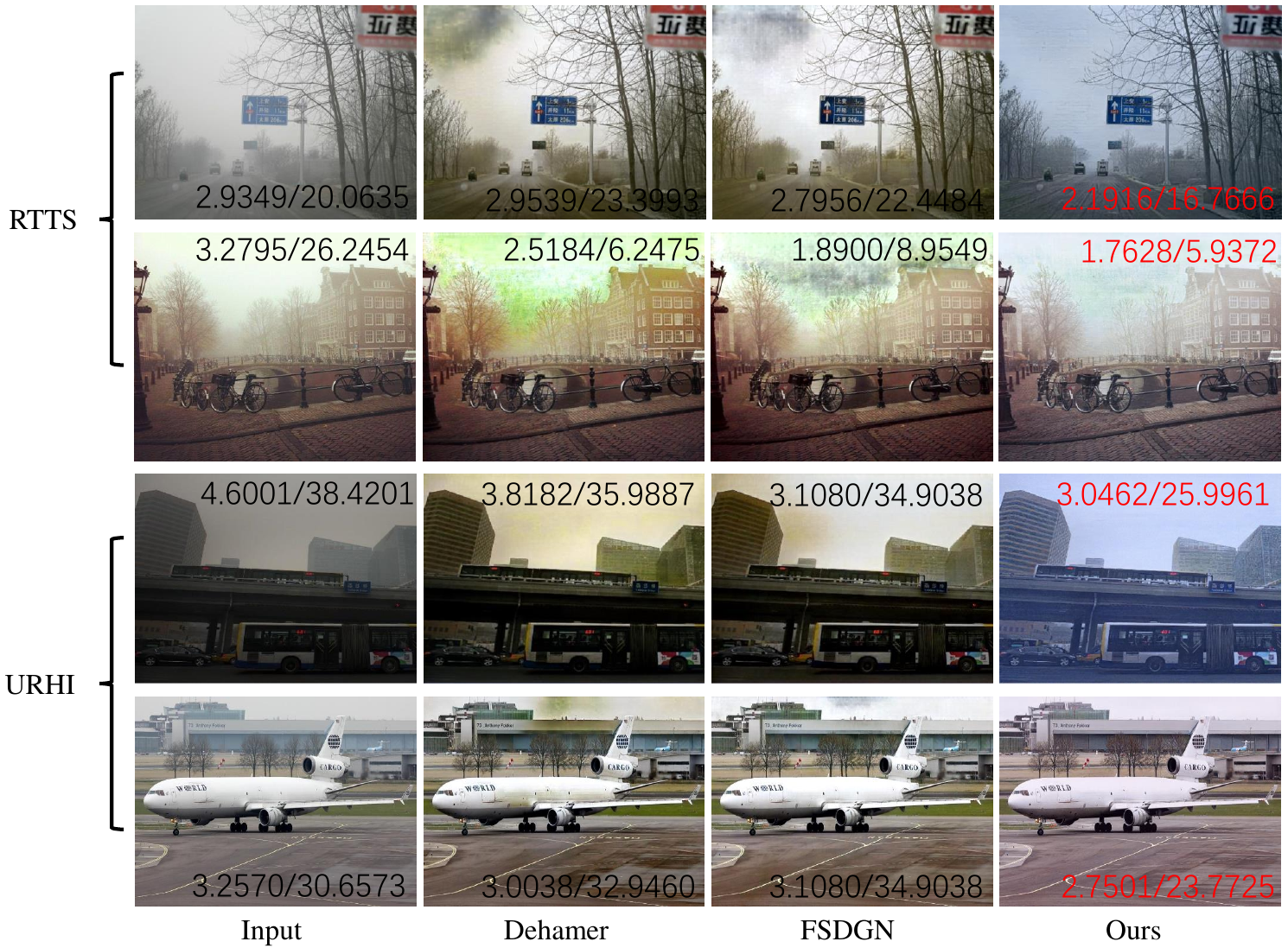}
	\caption{Visual results comparison on URHI and RTTS datasets. NIQE/BRISQUE values of per image are shown above the images.}
	\label{Visualization_URHI}
\end{figure}

Table~\ref{performance_Generalization} compares the quantitative results of different methods on URHI and RTTS datasets. Our method achieves the highest no-reference image quality assessment indicators, validating the superior generalization ability of our method compared to existing SOTA methods. We also show visual comparison in Fig.~\ref{Visualization_URHI}. Obviously, our method generates high-quality images with clearer architecture and finer details. 


\begin{table}[htbp]
\small
\setlength{\tabcolsep}{6pt}
\caption{Quantitative comparisons with SOTAs on the real-world URHI and RTTS datasets. $\downarrow$ denotes lower is better.}
\label{performance_Generalization}
\centering
\begin{tabular}{lcccc}
	\toprule
	\multirow{2}{*}{Method} & \multicolumn{2}{c}{\textbf{URHI}}& \multicolumn{2}{c}{\textbf{RTTS}} \\

	\cmidrule(lr){2-3} \cmidrule(lr){4-5}
	& BRISQUE$\downarrow$  & NIQE$\downarrow$  & BRISQUE$\downarrow$  & NIQE$\downarrow$  \\
	
	\midrule
	Dehamer   &  29.0822  & 3.8415 &  34.2643  & 4.4089   \\
	FSDGN     &  29.4439  & 3.8161 &  30.7974  & 4.1260  \\ 
	\midrule
	Ours      & \textbf{17.8439} & \textbf{3.4752} & \textbf{19.4593} & \textbf{3.5945}  \\ 
	\bottomrule
\end{tabular}
\end{table}

\noindent
{\bf GridNet as the backbone of the first stage.}
In this paper, we select the SOTA dehazing method FSDGN \cite{yufrequency} as the backbone of the first stage. Besides, we also employ the classical GridNet \cite{liu2019griddehazenet} as a weaker backbone of the first stage to validate the effectiveness and robustness of our method. 
Table~\ref{GridNet_Performance} show the quantitative results of our method. We can draw two points from this table: (1) Even with a weaker backbone for the first stage, our method still gets substantial performance improvement over the backbone, which shows the robustness and effectiveness of our method. (2) The performance of our method with GridNet as first stage's backbone is inferior to that with FSDGN as first stage's backbone. This verifies the first stage's backbone selection principle and the reasonability of our strategy pulling the distribution closer.

\begin{table}[htbp]
\caption{Quantitative comparisons with GridNet as backbone of the first stage on the NH-HAZE and Dense\_Haze dataset.}
\label{GridNet_Performance}
\setlength{\tabcolsep}{6pt}
\centering
\begin{tabular}{cc|cccc}
	\hline 
	Dataset & Method & FID$\downarrow$  & LPIPS$\downarrow$ & PSNR$\uparrow$ & SSIM$\uparrow$\\ \hline 
	\multirow{3}{*}{NH} &  Hazy   & 446.9583  & 0.4786 & 11.3375  & 0.4063 \\
        & First-stage & 224.0246  & 0.3152  & 17.6350  & 0.5929 \\
 	&  DehazeDDPM  &  \textbf{134.4730} & \textbf{0.1969} & \textbf{19.9503} & \textbf{0.5983}\\ 
 	\hline
        \multirow{3}{*}{Dense}  &  Hazy  & 442.4977  & 0.5936 & 8.5041  & 0.4243 \\
        &  First-stage & 393.5054  & 0.5197  & 13.4489  & 0.4777 \\
  	&  DehazeDDPM  &  \textbf{146.0102} & \textbf{0.3452} & \textbf{17.3070} & \textbf{0.4881}\\ 
	\hline
\end{tabular}
\end{table}

\begin{figure}[htbp]
	\begin{center}
		\includegraphics[width=0.985\linewidth]{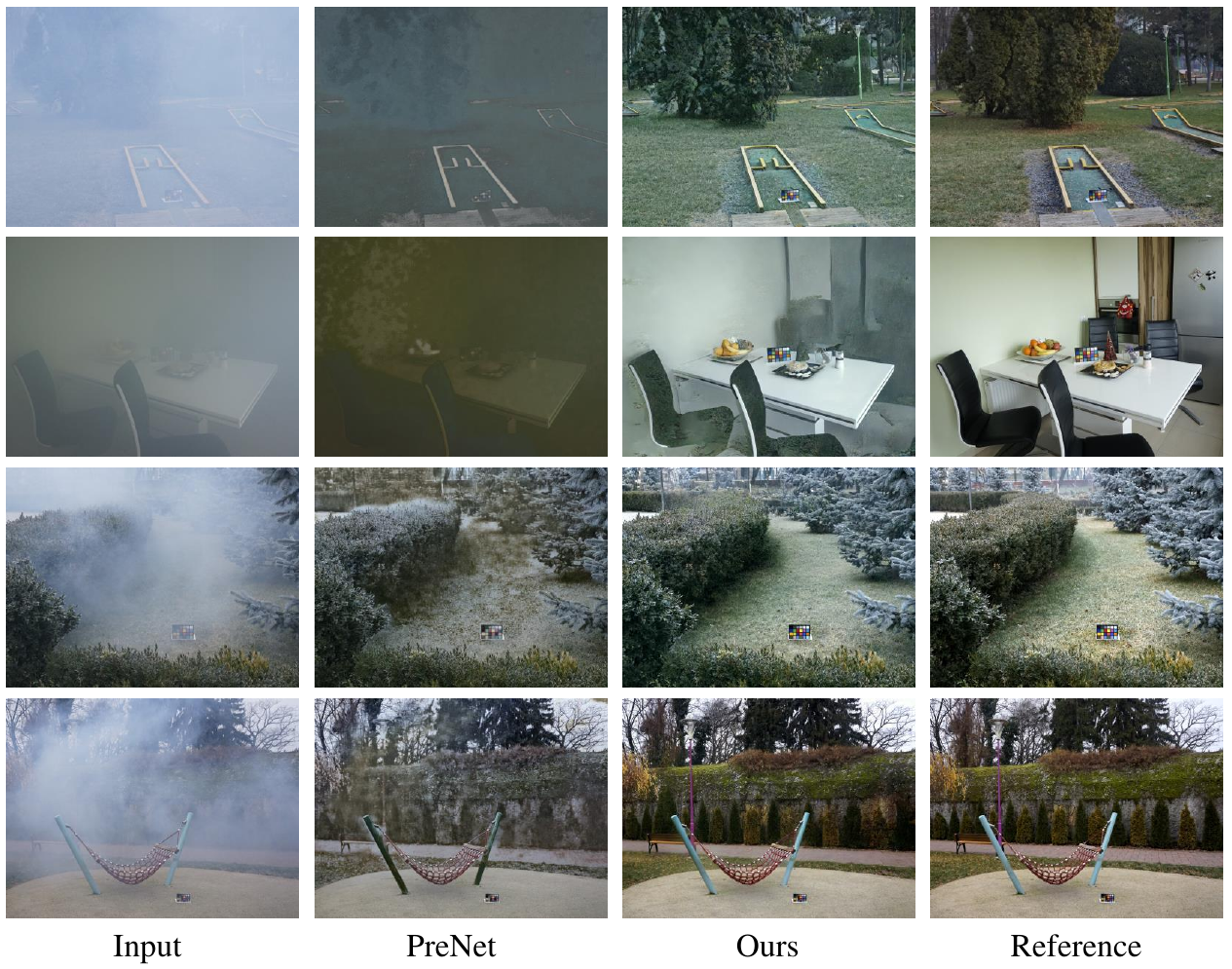}
	\end{center}
	\caption{Comparison of visual results on Dense-Haze and NH-HAZE dataset with GridNet \cite{liu2019griddehazenet} as the backbone of the first stage. The images in the first two rows are from Dense-Haze dataset and the bottom two are from NH-HAZE dataset.}
	\label{GridNet}
\end{figure}

\begin{figure}[htbp]
	\begin{center}
		\includegraphics[width=0.985\linewidth]{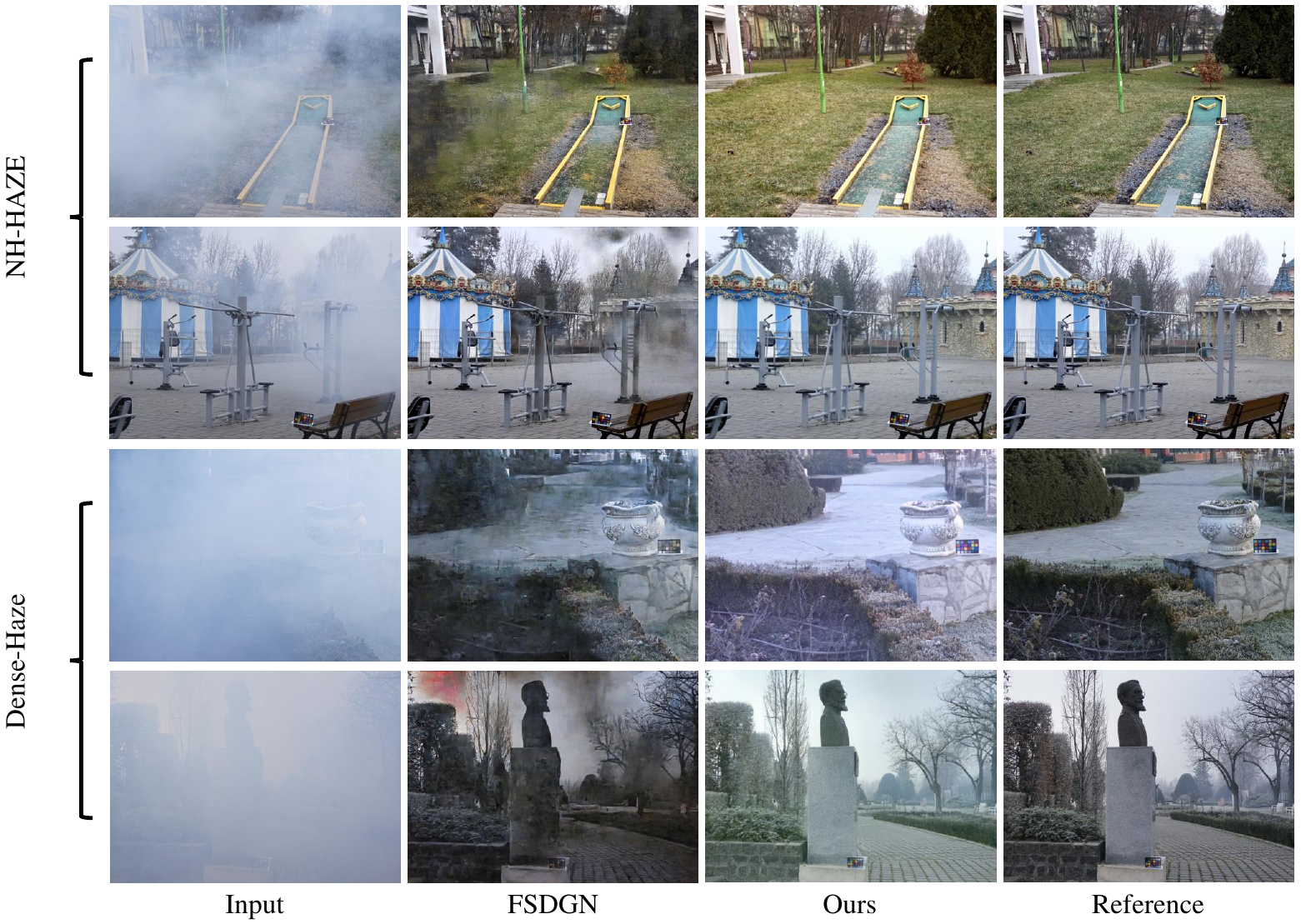}
	\end{center}
	\caption{Comparison of visual results for "seen" data on NH-HAZE and Dense-Haze datasets. Zoom in for best view.}
	\label{DENSE-NH}
\end{figure}

\noindent
{\bf Memorizing capacity for training dataset.}
Besides the above comparison, we also conduct more experiments to validate the superiority of DehazeDDPM over existing image dehazing methods on real hazy images. Concretely, we validate the dehazing performance of DehazeDDPM on the real hazy image falling in the training set. That is to say, we intend to figure out the performance of our method for the data that they have “seen" in the training stage. We show some examples in Fig. \ref{DENSE-NH}. 
Surprisingly, our DehazeDDPM recovers nearly all the original information. The recovered immense information is absent in the input image and only exists in the reference image, which verifies that our method can truly memorize the knowledge of clear image by learning its distribution. In contrast, previous mapping-based methods still performs poorly even if testing on the training dataset.

\subsection{Ablation Studies}
In this section, we perform several ablation studies to analyse the effectiveness of the proposed method on Dense-Haze dataset. The studies include the following ablated models:
(a) Joint training of the two stages: Instead of freezing the parameters of the first stage, we joint train the two stages in the training stage. 
(b)w/o Fog-aware and Distribution-closer Conditions: Instead of $trmap$ and $J$, we employ the original hazy image $I$ as condition, just as existing conditional DDPM does.
(c) w/o Confidence-guided Dynamic Fusion: We no not use the CDF in the denoise process. 
(d) w/o Frequency prior loss: We no not use the Frequency prior loss in the training stage. 
(e) Ours(full model): Our final setting in this method. 
These models are trained using the same training setting as our method. The performance of these models are summarized in Table~\ref{tab:ablation}. Obviously, our two-stage strategy, FDC, CDF and frequency loss all elevate the performance.

\begin{table}[htbp]
\small
\setlength{\tabcolsep}{11pt}
\caption{The results of ablated models on Dense-Haze \cite{Dense-Haze_2019} dataset.}
\label{tab:ablation}
\centering
\begin{tabular}{lcccc}
	\toprule
	\multirow{2}{*}{Label} & \multicolumn{2}{c}{\textbf{Perceptual}}& \multicolumn{2}{c}{\textbf{Distortion}} \\

	\cmidrule(lr){2-3} \cmidrule(lr){4-5}
	& FID$\downarrow$ & LPIPS$\downarrow$  & PSNR$\uparrow$ & SSIM$\uparrow$  \\
	
	\midrule
    a  & 181.24 & 0.3323 & 18.21 & 0.5786   \\
    b  & 194.28 & 0.3639 & 17.86 & 0.5704   \\
    c  & 178.63 & 0.3112 & 18.58 & 0.5832   \\
    d  & 174.06 & 0.2947 & 18.81 & 0.5901   \\
    e  & 167.32 & 0.3022 & 19.14 & 0.5922   \\
	\bottomrule
\end{tabular}
\end{table}

\section{Discussion and future work}
\label{sec:Discussion}
Diffusion models can memorize the distribution of data, which means that a larger data size may help more. In our experiment, the Dense-Haze and NH-HAZE both are small scale datasets, somewhat restricting the strong generation ability of Diffusion models in image dehazing. A real-world large scale image dehazing dataset is necessary for our future work as well as the further development of dehazing community. While, although with small scale dataset, our method still demonstrates unprecedented perceptual quality in image dehazing task. It worth noting that our method generates more natural results than existing methods. Our work views image dehazing task from the deep generative modelling perspective and demonstrates unprecedented perceptual quality in image dehazing task. Our method provides a promising solution for complex real-world image dehazing task.


\section{Conclusion}
\label{sec:conclusion}
In this paper, we propose DehazeDDPM: A DDPM-based and physics-aware image dehazing framework that is applicable to complex hazy scenarios. Our DehazeDDPM firstly introduce conditional DDPM to tackle the challenging dense-haze image dehazing task by working in conjunction with the physical modelling. Specifically, DehazeDDPM works in two stages. The former physical modelling stage pulls the distribution of hazy data closer to that of clear data and endows DehazeDDPM with fog-aware ability. The latter stage exploits the strong generation ability of DDPM to recover the haze-induced information loss. Extensive experiments demonstrate that our method attains SOTA performance on several image dehazing benchmarks.

\bibliographystyle{IEEEtran}
\bibliography{main}

\begin{thebibliography}{10}
\providecommand{\url}[1]{#1}
\csname url@samestyle\endcsname
\providecommand{\newblock}{\relax}
\providecommand{\bibinfo}[2]{#2}
\providecommand{\BIBentrySTDinterwordspacing}{\spaceskip=0pt\relax}
\providecommand{\BIBentryALTinterwordstretchfactor}{4}
\providecommand{\BIBentryALTinterwordspacing}{\spaceskip=\fontdimen2\font plus
\BIBentryALTinterwordstretchfactor\fontdimen3\font minus \fontdimen4\font\relax}
\providecommand{\BIBforeignlanguage}[2]{{%
\expandafter\ifx\csname l@#1\endcsname\relax
\typeout{** WARNING: IEEEtran.bst: No hyphenation pattern has been}%
\typeout{** loaded for the language `#1'. Using the pattern for}%
\typeout{** the default language instead.}%
\else
\language=\csname l@#1\endcsname
\fi
#2}}
\providecommand{\BIBdecl}{\relax}
\BIBdecl

\bibitem{mccartney1976optics}
E.~J. McCartney, ``Optics of the atmosphere: scattering by molecules and particles,'' \emph{New York}, 1976.

\bibitem{narasimhan2002vision}
S.~G. Narasimhan and S.~K. Nayar, ``Vision and the atmosphere,'' \emph{International journal of computer vision}, vol.~48, no.~3, pp. 233--254, 2002.

\bibitem{guo2022image}
C.-L. Guo, Q.~Yan, S.~Anwar, R.~Cong, W.~Ren, and C.~Li, ``Image dehazing transformer with transmission-aware 3d position embedding,'' in \emph{Proceedings of the IEEE/CVF Conference on Computer Vision and Pattern Recognition}, 2022, pp. 5812--5820.

\bibitem{berman2016non}
D.~Berman, S.~Avidan \emph{et~al.}, ``Non-local image dehazing,'' in \emph{Proceedings of the IEEE Conference on Computer Vision and Pattern Recognition}, 2016, pp. 1674--1682.

\bibitem{he2010single}
K.~He, J.~Sun, and X.~Tang, ``Single image haze removal using dark channel prior,'' \emph{IEEE Transactions on Pattern Analysis and Machine Intelligence}, vol.~33, no.~12, pp. 2341--2353, 2010.

\bibitem{fattal2014dehazing}
R.~Fattal, ``Dehazing using color-lines,'' \emph{ACM Transactions on Graphics (TOG)}, vol.~34, no.~1, pp. 1--14, 2014.

\bibitem{fattal2008single}
------, ``Single image dehazing,'' \emph{ACM Transactions on Graphics (TOG)}, vol.~27, no.~3, pp. 1--9, 2008.

\bibitem{li2017aod}
B.~Li, X.~Peng, Z.~Wang, J.~Xu, and D.~Feng, ``{AOD-Net}: All-in-one dehazing network,'' in \emph{Proceedings of the IEEE International Conference on Computer Vision}, 2017, pp. 4770--4778.

\bibitem{liu2019griddehazenet}
X.~Liu, Y.~Ma, Z.~Shi, and J.~Chen, ``{GridDehazeNet}: Attention-based multi-scale network for image dehazing,'' in \emph{Proceedings of the IEEE/CVF International Conference on Computer Vision}, 2019, pp. 7314--7323.

\bibitem{guo2019dense}
T.~Guo, X.~Li, V.~Cherukuri, and V.~Monga, ``Dense scene information estimation network for dehazing,'' in \emph{Proceedings of the IEEE/CVF Conference on Computer Vision and Pattern Recognition Workshops}, 2019, pp. 0--0.

\bibitem{dong2020multi}
H.~Dong, J.~Pan, L.~Xiang, Z.~Hu, X.~Zhang, F.~Wang, and M.-H. Yang, ``Multi-scale boosted dehazing network with dense feature fusion,'' in \emph{Proceedings of the IEEE/CVF Conference on Computer Vision and Pattern Recognition}, 2020, pp. 2157--2167.

\bibitem{wu2021contrastive}
H.~Wu, Y.~Qu, S.~Lin, J.~Zhou, R.~Qiao, Z.~Zhang, Y.~Xie, and L.~Ma, ``Contrastive learning for compact single image dehazing,'' in \emph{Proceedings of the IEEE/CVF Conference on Computer Vision and Pattern Recognition}, 2021, pp. 10\,551--10\,560.

\bibitem{yufrequency}
H.~Yu, N.~Zheng, M.~Zhou, J.~Huang, Z.~Xiao, and F.~Zhao, ``Frequency and spatial dual guidance for image dehazing,'' 2022.

\bibitem{van2008visualizing}
L.~Van~der Maaten and G.~Hinton, ``Visualizing data using t-sne.'' \emph{Journal of machine learning research}, vol.~9, no.~11, 2008.

\bibitem{sohl2015deep}
J.~Sohl-Dickstein, E.~Weiss, N.~Maheswaranathan, and S.~Ganguli, ``Deep unsupervised learning using nonequilibrium thermodynamics,'' in \emph{International Conference on Machine Learning}.\hskip 1em plus 0.5em minus 0.4em\relax PMLR, 2015, pp. 2256--2265.

\bibitem{song2019generative}
Y.~Song and S.~Ermon, ``Generative modeling by estimating gradients of the data distribution,'' \emph{Advances in Neural Information Processing Systems}, vol.~32, 2019.

\bibitem{ho2020denoising}
J.~Ho, A.~Jain, and P.~Abbeel, ``Denoising diffusion probabilistic models,'' \emph{Advances in Neural Information Processing Systems}, vol.~33, pp. 6840--6851, 2020.

\bibitem{nichol2021improved}
A.~Q. Nichol and P.~Dhariwal, ``Improved denoising diffusion probabilistic models,'' in \emph{International Conference on Machine Learning}.\hskip 1em plus 0.5em minus 0.4em\relax PMLR, 2021, pp. 8162--8171.

\bibitem{dhariwal2021diffusion}
P.~Dhariwal and A.~Nichol, ``Diffusion models beat gans on image synthesis,'' \emph{Advances in Neural Information Processing Systems}, vol.~34, pp. 8780--8794, 2021.

\bibitem{saharia2022image}
C.~Saharia, J.~Ho, W.~Chan, T.~Salimans, D.~J. Fleet, and M.~Norouzi, ``Image super-resolution via iterative refinement,'' \emph{IEEE Transactions on Pattern Analysis and Machine Intelligence}, 2022.

\bibitem{whang2022deblurring}
J.~Whang, M.~Delbracio, H.~Talebi, C.~Saharia, A.~G. Dimakis, and P.~Milanfar, ``Deblurring via stochastic refinement,'' in \emph{Proceedings of the IEEE/CVF Conference on Computer Vision and Pattern Recognition}, 2022, pp. 16\,293--16\,303.

\bibitem{rombach2022high}
R.~Rombach, A.~Blattmann, D.~Lorenz, P.~Esser, and B.~Ommer, ``High-resolution image synthesis with latent diffusion models,'' in \emph{Proceedings of the IEEE/CVF Conference on Computer Vision and Pattern Recognition}, 2022, pp. 10\,684--10\,695.

\bibitem{zhu2014single}
Q.~Zhu, J.~Mai, and L.~Shao, ``Single image dehazing using color attenuation prior.'' in \emph{BMVC}.\hskip 1em plus 0.5em minus 0.4em\relax Citeseer, 2014.

\bibitem{chen2016robust}
C.~Chen, M.~N. Do, and J.~Wang, ``Robust image and video dehazing with visual artifact suppression via gradient residual minimization,'' in \emph{Proceedings of the European Conference on Computer Vision}.\hskip 1em plus 0.5em minus 0.4em\relax Springer, 2016, pp. 576--591.

\bibitem{zhang2017fast}
J.~Zhang, Y.~Cao, S.~Fang, Y.~Kang, and C.~Wen~Chen, ``Fast haze removal for nighttime image using maximum reflectance prior,'' in \emph{Proceedings of the IEEE Conference on Computer Vision and Pattern Recognition}, 2017, pp. 7418--7426.

\bibitem{cai2016dehazenet}
B.~Cai, X.~Xu, K.~Jia, C.~Qing, and D.~Tao, ``{DehazeNet}: An end-to-end system for single image haze removal,'' \emph{IEEE Transactions on Image Processing}, vol.~25, no.~11, pp. 5187--5198, 2016.

\bibitem{ren2016single}
W.~Ren, S.~Liu, H.~Zhang, J.~Pan, X.~Cao, and M.-H. Yang, ``Single image dehazing via multi-scale convolutional neural networks,'' in \emph{Proceedings of the European Conference on Computer Vision}.\hskip 1em plus 0.5em minus 0.4em\relax Springer, 2016, pp. 154--169.

\bibitem{zhang2018densely}
H.~Zhang and V.~M. Patel, ``Densely connected pyramid dehazing network,'' in \emph{Proceedings of the IEEE Conference on Computer Vision and Pattern Recognition}, 2018, pp. 3194--3203.

\bibitem{liu2018learning}
R.~Liu, X.~Fan, M.~Hou, Z.~Jiang, Z.~Luo, and L.~Zhang, ``Learning aggregated transmission propagation networks for haze removal and beyond,'' \emph{IEEE transactions on neural networks and learning systems}, vol.~30, no.~10, pp. 2973--2986, 2018.

\bibitem{liu2019learning}
Y.~Liu, J.~Pan, J.~Ren, and Z.~Su, ``Learning deep priors for image dehazing,'' in \emph{Proceedings of the IEEE/CVF International Conference on Computer Vision}, 2019, pp. 2492--2500.

\bibitem{li2016underwater}
C.~Li, J.~Guo, R.~Cong, Y.~Pang, and B.~Wang, ``Underwater image enhancement by dehazing with minimum information loss and histogram distribution prior,'' \emph{IEEE Transactions on Image Processing}, vol.~25, no.~12, pp. 5664--5677, 2016.

\bibitem{ren2018gated}
W.~Ren, L.~Ma, J.~Zhang, J.~Pan, X.~Cao, W.~Liu, and M.-H. Yang, ``Gated fusion network for single image dehazing,'' in \emph{Proceedings of the IEEE Conference on Computer Vision and Pattern Recognition}, 2018, pp. 3253--3261.

\bibitem{deng2020hardgan}
Q.~Deng, Z.~Huang, C.-C. Tsai, and C.-W. Lin, ``{HardGAN}: A haze-aware representation distillation gan for single image dehazing,'' in \emph{Proceedings of the European Conference on Computer Vision}.\hskip 1em plus 0.5em minus 0.4em\relax Springer, 2020, pp. 722--738.

\bibitem{qin2020ffa}
X.~Qin, Z.~Wang, Y.~Bai, X.~Xie, and H.~Jia, ``{FFA-Net}: Feature fusion attention network for single image dehazing,'' in \emph{Proceedings of the AAAI Conference on Artificial Intelligence}, vol.~34, 2020, pp. 11\,908--11\,915.

\bibitem{DehazeYu}
H.~Yu, J.~Huang, Y.~Liu, Q.~Zhu, M.~Zhou, and F.~Zhao, ``Source-free domain adaptation for real-world image dehazing,'' in \emph{Proceedings of the 30th ACM International Conference on Multimedia}, 2022, p. 6645–6654.

\bibitem{liu2022towards}
H.~Liu, Z.~Wu, L.~Li, S.~Salehkalaibar, J.~Chen, and K.~Wang, ``Towards multi-domain single image dehazing via test-time training,'' in \emph{Proceedings of the IEEE/CVF Conference on Computer Vision and Pattern Recognition}, 2022, pp. 5831--5840.

\bibitem{zhou2022fsad}
Y.~Zhou, Z.~Chen, P.~Li, H.~Song, C.~P. Chen, and B.~Sheng, ``Fsad-net: feedback spatial attention dehazing network,'' \emph{IEEE transactions on neural networks and learning systems}, 2022.

\bibitem{fan2022multiscale}
G.~Fan, M.~Gan, B.~Fan, and C.~P. Chen, ``Multiscale cross-connected dehazing network with scene depth fusion,'' \emph{IEEE Transactions on Neural Networks and Learning Systems}, 2022.

\bibitem{zheng2023curricular}
Y.~Zheng, J.~Zhan, S.~He, J.~Dong, and Y.~Du, ``Curricular contrastive regularization for physics-aware single image dehazing,'' in \emph{Proceedings of the IEEE/CVF Conference on Computer Vision and Pattern Recognition}, 2023, pp. 5785--5794.

\bibitem{wu2023ridcp}
R.-Q. Wu, Z.-P. Duan, C.-L. Guo, Z.~Chai, and C.~Li, ``Ridcp: Revitalizing real image dehazing via high-quality codebook priors,'' in \emph{Proceedings of the IEEE/CVF Conference on Computer Vision and Pattern Recognition}, 2023, pp. 22\,282--22\,291.

\bibitem{van2016conditional}
A.~Van~den Oord, N.~Kalchbrenner, L.~Espeholt, O.~Vinyals, A.~Graves \emph{et~al.}, ``Conditional image generation with pixelcnn decoders,'' \emph{Advances in neural information processing systems}, vol.~29, 2016.

\bibitem{kingma2013auto}
D.~P. Kingma and M.~Welling, ``Auto-encoding variational bayes,'' \emph{arXiv preprint arXiv:1312.6114}, 2013.

\bibitem{kingma2018glow}
D.~P. Kingma and P.~Dhariwal, ``Glow: Generative flow with invertible 1x1 convolutions,'' \emph{Advances in neural information processing systems}, vol.~31, 2018.

\bibitem{goodfellow2020generative}
I.~Goodfellow, J.~Pouget-Abadie, M.~Mirza, B.~Xu, D.~Warde-Farley, S.~Ozair, A.~Courville, and Y.~Bengio, ``Generative adversarial networks,'' \emph{Advances in Neural Information Processing Systems}, vol.~63, 2014.

\bibitem{li2021dehazeflow}
H.~Li, J.~Li, D.~Zhao, and L.~Xu, ``Dehazeflow: Multi-scale conditional flow network for single image dehazing,'' in \emph{Proceedings of the 29th ACM International Conference on Multimedia}, 2021, pp. 2577--2585.

\bibitem{dong2020fd}
Y.~Dong, Y.~Liu, H.~Zhang, S.~Chen, and Y.~Qiao, ``Fd-gan: Generative adversarial networks with fusion-discriminator for single image dehazing,'' in \emph{Proceedings of the AAAI Conference on Artificial Intelligence}, vol.~34, 2020, pp. 10\,729--10\,736.

\bibitem{fu2021dw}
M.~Fu, H.~Liu, Y.~Yu, J.~Chen, and K.~Wang, ``{DW-GAN}: A discrete wavelet transform gan for nonhomogeneous dehazing,'' in \emph{Proceedings of the IEEE/CVF Conference on Computer Vision and Pattern Recognition}, 2021, pp. 203--212.

\bibitem{sharma2020scale}
P.~Sharma, P.~Jain, and A.~Sur, ``Scale-aware conditional generative adversarial network for image dehazing,'' in \emph{Proceedings of the IEEE/CVF Winter Conference on Applications of Computer Vision}, 2020, pp. 2355--2365.

\bibitem{kumar2022orthogonal}
A.~Kumar, M.~Sanathra, M.~Khare, and V.~Khare, ``Orthogonal transform based generative adversarial network for image dehazing,'' \emph{arXiv preprint arXiv:2206.01743}, 2022.

\bibitem{razavi2019generating}
A.~Razavi, A.~Van~den Oord, and O.~Vinyals, ``Generating diverse high-fidelity images with vq-vae-2,'' \emph{Advances in neural information processing systems}, vol.~32, 2019.

\bibitem{gulrajani2017improved}
I.~Gulrajani, F.~Ahmed, M.~Arjovsky, V.~Dumoulin, and A.~C. Courville, ``Improved training of wasserstein gans,'' \emph{Advances in neural information processing systems}, vol.~30, 2017.

\bibitem{miyato2018spectral}
T.~Miyato, T.~Kataoka, M.~Koyama, and Y.~Yoshida, ``Spectral normalization for generative adversarial networks,'' \emph{arXiv preprint arXiv:1802.05957}, 2018.

\bibitem{choi2021ilvr}
J.~Choi, S.~Kim, Y.~Jeong, Y.~Gwon, and S.~Yoon, ``Ilvr: Conditioning method for denoising diffusion probabilistic models,'' \emph{arXiv preprint arXiv:2108.02938}, 2021.

\bibitem{lugmayr2022repaint}
A.~Lugmayr, M.~Danelljan, A.~Romero, F.~Yu, R.~Timofte, and L.~Van~Gool, ``Repaint: Inpainting using denoising diffusion probabilistic models,'' in \emph{Proceedings of the IEEE/CVF Conference on Computer Vision and Pattern Recognition}, 2022, pp. 11\,461--11\,471.

\bibitem{ozdenizci2022restoring}
O.~{\"O}zdenizci and R.~Legenstein, ``Restoring vision in adverse weather conditions with patch-based denoising diffusion models,'' \emph{arXiv preprint arXiv:2207.14626}, 2022.

\bibitem{chen2021psd}
Z.~Chen, Y.~Wang, Y.~Yang, and D.~Liu, ``{PSD}: Principled synthetic-to-real dehazing guided by physical priors,'' in \emph{Proceedings of the IEEE/CVF conference on computer vision and pattern recognition}, 2021, pp. 7180--7189.

\bibitem{li2018benchmarking}
B.~Li, W.~Ren, D.~Fu, D.~Tao, D.~Feng, W.~Zeng, and Z.~Wang, ``Benchmarking single-image dehazing and beyond,'' \emph{IEEE Transactions on Image Processing}, vol.~28, no.~1, pp. 492--505, 2018.

\bibitem{Dense-Haze_2019}
C.~O. Ancuti, C.~Ancuti, M.~Sbert, and R.~Timofte, ``Dense haze: A benchmark for image dehazing with dense-haze and haze-free images,'' in \emph{Proceedings of the IEEE International Conference on Image Processing}, 2019, pp. 1014--1018.

\bibitem{ancuti2020nh}
C.~O. Ancuti, C.~Ancuti, and R.~Timofte, ``{NH-HAZE}: An image dehazing benchmark with non-homogeneous hazy and haze-free images,'' in \emph{Proceedings of the IEEE/CVF Conference on Computer Vision and Pattern Recognition Workshops}, 2020, pp. 444--445.

\bibitem{song2020improved}
Y.~Song and S.~Ermon, ``Improved techniques for training score-based generative models,'' \emph{Advances in neural information processing systems}, vol.~33, pp. 12\,438--12\,448, 2020.

\bibitem{deck2023easing}
K.~Deck and T.~Bischoff, ``Easing color shifts in score-based diffusion models,'' \emph{arXiv preprint arXiv:2306.15832}, 2023.

\bibitem{he2019bag}
T.~He, Z.~Zhang, H.~Zhang, Z.~Zhang, J.~Xie, and M.~Li, ``Bag of tricks for image classification with convolutional neural networks,'' in \emph{Proceedings of the IEEE/CVF Conference on Computer Vision and Pattern Recognition}, 2019, pp. 558--567.

\bibitem{heusel2017gans}
M.~Heusel, H.~Ramsauer, T.~Unterthiner, B.~Nessler, and S.~Hochreiter, ``Gans trained by a two time-scale update rule converge to a local nash equilibrium,'' \emph{Advances in neural information processing systems}, vol.~30, 2017.

\bibitem{zhang2018unreasonable}
R.~Zhang, P.~Isola, A.~A. Efros, E.~Shechtman, and O.~Wang, ``The unreasonable effectiveness of deep features as a perceptual metric,'' in \emph{Proceedings of the IEEE conference on computer vision and pattern recognition}, 2018, pp. 586--595.

\bibitem{wang2004image}
Z.~Wang, A.~C. Bovik, H.~R. Sheikh, and E.~P. Simoncelli, ``Image quality assessment: from error visibility to structural similarity,'' \emph{IEEE transactions on image processing}, vol.~13, pp. 600--612, 2004.

\bibitem{mittal2012no}
A.~Mittal, A.~K. Moorthy, and A.~C. Bovik, ``No-reference image quality assessment in the spatial domain,'' \emph{IEEE Transactions on image processing}, vol.~21, no.~12, pp. 4695--4708, 2012.

\bibitem{mittal2012making}
A.~Mittal, R.~Soundararajan, and A.~C. Bovik, ``Making a “completely blind” image quality analyzer,'' \emph{IEEE Signal processing letters}, vol.~20, no.~3, pp. 209--212, 2012.

\end{thebibliography}

\end{document}